\documentclass[11pt,a4paper]{article}       

\usepackage[T1]{fontenc}
\usepackage[utf8]{inputenc}
\usepackage{authblk}
\usepackage{graphicx}
\usepackage{amssymb}
\usepackage{amsmath}
\usepackage{enumerate}
\usepackage{subfigure}
\usepackage{url,ulem,rotating,tabularx,color}
\usepackage{algorithm}
\usepackage{algorithmicx}
\newcommand{\ba}{\begin{array}}
\newcommand{\ea}{\end{array}}

\newcommand{\bc}{\begin{center}}
\newcommand{\ec}{\end{center}}
\newcommand{\bit}{\begin{itemize}}
\newcommand{\eit}{\end{itemize}}

\newcommand{\beq}{\begin{equation}}
\newcommand{\eeq}{\end{equation}}

\newcommand{\bfp}{\mathbf{p}}
\newcommand{\bfq}{\mathbf{q}}

\newcommand{\bfx}{\mathbf{x}}

\newcommand{\bfK}{\mathbf{K}}

\newcommand{\bfM}{\mathbf{M}}

\newcommand{\bfP}{\mathbf{P}}

\newcommand{\bfR}{\mathbf{R}}

\newcommand{\bfT}{\mathbf{T}}

\newcommand{\be}{\begin{equation}}
\newcommand{\ee}{\end{equation}}
\newcommand{\beqa}{\begin{eqnarray}}
\newcommand{\eeqa}{\end{eqnarray}}

\begin{document}

\title{Noise in Structured-Light Stereo Depth Cameras: Modeling and its Applications}
\author[1]{Avishek Chatterjee\thanks{avishek@ee.iisc.ernet.in}}
\author[1]{Venu Madhav Govindu\thanks{venu@ee.iisc.ernet.in}}
\affil[1]{Dept. of Electrical Engineering, Indian Institute of Science, Bengaluru -560012}

\date{}

\maketitle

\begin{abstract}
Depth maps obtained from commercially available structured-light stereo based depth cameras, such as the Kinect, are easy to use but are affected by significant amounts of noise. This paper is devoted to a study of the intrinsic noise characteristics of such depth maps, i.e. the standard deviation of noise in estimated depth varies quadratically with the distance of the object from the depth camera. We validate this theoretical model against empirical observations and demonstrate the utility of this noise model in three popular applications: depth map denoising, volumetric scan merging for 3D modeling, and identification of 3D planes in depth maps.

\end{abstract}

\section{Introduction}\label{sec:introduction}
3D scanning is used extensively for many computer vision applications, e.g. human-computer interface (HCI), virtual reality, game programming, industrial monitoring, archeology, etc. Although applications such as HCI or gaming generally demand speed rather than precision, the accuracy of the 3D reconstruction is of crucial importance for many other tasks including archeology or quality monitoring in industrial production. The recent commercial availability of inexpensive structured-light depth cameras has opened up new possibilities for 3D scanning and shape reconstruction. Such devices were originally intended for human pose estimation in a gaming context but an extensive body of research by the vision community has demonstrated their effective use in 3D shape reconstruction \cite{RGBDMapping,KinectFusion,EfficientKinfu,Kintinuous,MovingKinfu,HashingReconstruction,ScalableReconstruction,CPU3dMapping,InterestPoints,ElasticFragments}. While the low cost, ease of use and availability of depth maps at video frame rate make such depth cameras very attractive, these devices do suffer from high level of noise in the raw depth maps which needs to be addressed before such depth cameras can be used for 3D scanning or reconstruction. Therefore, it is crucial to understand the accuracy and noise characteristics of structured-light depth cameras and develop appropriate methods to mitigate the effects of such noise on the final 3D reconstructions or other representations. In the remainder of this section we briefly survey the current literature in this regard. In Section \ref{Section:Noisecharacteristic}, we briefly discuss the working principle of a structured-light stereo based depth camera and show how the quadratic nature of depth noise is inherent to the working principle of such depth cameras. We also validate our claim with thorough experimentation. In Section \ref{Section:Applications}, we show the effectiveness of this noise modeling in depth map denoising, volumetric scan merging and plane extraction. 
Finally, we provide some conclusions in Section \ref{Sec:Conclusion}.

The accuracy and noise characteristics of structured-light depth cameras like the Kinect has been investigated in recent years \cite{chatterjee2012pipeline,nguyen2012modeling,smisek20133d,khoshelham2011accuracy,khoshelham2012accuracy}. A good description of the working principle of the Kinect is available in \cite{martinezkinect,KinectROSSpec}. In addition, several studies have been performed to denoise the depth maps obtained from such devices. We may categorize these approaches into ones that make no explicit attempt to characterize the depth noise \cite{KinectFusion,milani2012joint,camplani2013depth,qi2013structure,yu2013shading,hanhigh,WangZPQ14,LiuGL12,LoWH13,KimCKA11,garcia2013real,ChenLL12} and those that study the behavior of noise in the depth images \cite{nguyen2012modeling,smisek20133d,khoshelham2011accuracy,khoshelham2012accuracy,chatterjee2012pipeline}. Amongst the methods of the first category that do not model depth noise, a number of applications, including Kinect Fusion \cite{KinectFusion}, use bilateral smoothing of depth maps in their pipelines. Other works utilize the available RGB images captured simultaneously with the depth maps \cite{milani2012joint,camplani2013depth,qi2013structure}. In \cite{milani2012joint}, the authors exploit the assumption that edges in depth map and RGB image should occur together. Similar ideas are also exploited in \cite{camplani2013depth,qi2013structure,WangZPQ14,LiuGL12,LoWH13,KimCKA11,garcia2013real,ChenLL12}. In \cite {yu2013shading,hanhigh}, photometry is used to refine depth maps. However, \cite{KinectFusion,milani2012joint,camplani2013depth,qi2013structure,yu2013shading,hanhigh} do not recognize the underlying characteristic of noise inherent to structured-light stereo based depth cameras. Therefore, some of them rely on additional cues from RGB images to denoise depth maps.

In contrast, there are some other approaches that seek to understand the nature of noise present in depth camera. \cite{nguyen2012modeling} uses extensive empirical measurements to model the noise characteristic of depth sensors. They argue that accounting for the noise in this fashion significantly improves both reconstruction and tracking in the Kinect Fusion pipeline. Like \cite{nguyen2012modeling}, the work in \cite{smisek20133d} also empirically notes that the noise in Kinect depth maps have a quadratic relationship to depth. However, neither of these studies recognizes the fact that the quadratic nature of the standard-deviation of depth noise is inherent to the working principle of structured-light stereo based depth cameras. In \cite{khoshelham2011accuracy,khoshelham2012accuracy}, the working principle of structured-light stereo cameras was utilized to develop a model for the noise present in depth maps. However, no attempt was made in \cite{khoshelham2011accuracy,khoshelham2012accuracy} to incorporate the noise model into any application method or estimation framework.

In our work, we both derive a noise model for structured-light depth cameras and demonstrate its utility in a variety of applications. In an earlier conference paper~\cite{chatterjee2012pipeline} we independently derived the theoretical noise model like the one presented in~\cite{khoshelham2011accuracy,khoshelham2012accuracy}. In Section~\ref{Section:Noisecharacteristic}, we show that the quadratic nature of depth noise is inherent to the working principle of structured-light stereo depth cameras. This theoretical observation is also validated thorough experimental observation. Furthermore, unlike the work of \cite{khoshelham2011accuracy,khoshelham2012accuracy}, we demonstrate in Section~\ref{Section:Applications} how our theoretical understanding of depth map noise can be carefully exploited in a variety of applications such as depth map denoising, volumetric scan merging and 3D planar surface extraction. Before developing our noise model, we would like to comment on the nature of the present work. There exists a large body of literature on the ingredients of 3D reconstruction pipelines using depth cameras \cite{RGBDMapping,KinectFusion,EfficientKinfu,Kintinuous,MovingKinfu,HashingReconstruction,ScalableReconstruction,CPU3dMapping,InterestPoints,ElasticFragments}. Although we  also incorporate our noise model into a reconstruction pipeline to generate the 3D reconstruction results, this paper is not about scan registration or 3D reconstruction pipeline. Rather we emphasise the development of a noise model for depth cameras and demonstrate the value in using this noise model in various applications.

\section{A noise model for depth cameras}\label{Section:Noisecharacteristic}
In this Section, we briefly state the working principles of structured-light depth cameras and then theoretically derive a noise model for such depth cameras. We also provide empirical validation of the noise model that we develop. We use the Kinect as our depth camera for all the experiments in this section, although our ideas can also be applied to any other structured-light depth camera. The Kinect depth camera is a structured-light stereo system that consists of an infra-red (IR) projector and an infra-red camera. The projector is an IR laser operating at a wavelength of around 830nm that is placed behind a diffraction grating. This diffraction grating converts the single beam of the laser source into a fixed dot pattern. This pattern is projected on the 3D scene in front of the device and is reflected back to the IR camera of the device. Together the IR projector and camera act as a stereo pair. It may be noted here that a projector is geometrically equivalent to a pin-hole camera except that the direction of light is reversed. In addition, we recognise that the projector is equivalent to a bundle of rays with a common point. Therefore, in terms of projective geometry, we can associate a virtual projector plane where each projector pixel is a member of an equivalence class defined by a unique IR ray passing through it, i.e. projector pixels are the inhomogeneous representations of the corresponding IR rays. The reader is referred to \cite{HartleyZisserman} for details on the projective equivalence of pixels on a camera plane and a bundle of rays with a common point. From the above discussion we note that, for our geometrical analysis we can treat the Kinect as a pair of stereo cameras in canonical position.
\subsection{Noise Model}\label{Subsection:NoiseModel}
By correlating patches of the viewed IR camera with a known emitted pattern (i.e. known virtual projector image), one can obtain the disparity estimates in the IR camera plane~\cite{martinezkinect,KinectROSSpec}. The depth $(Z)$ or distance of an object is inversely proportional to this disparity $(D)$ i.e. $D = \frac{fB}{Z}$ where $B$ is the baseline distance between the optical centers of the camera and the projector and $f$ is the focal length of the camera. It has been observed that the Kinect is precise enough to need no stereo rectification. Similarly, non-linear lens distortions are negligible.

While factors such as sensor noise, the IR component of ambient light, reflectance of objects etc. cause inaccuracies in disparity measurement obtained with patch correlation, the most significant source of disparity error is quantization noise. Such quantization noise arises when we estimate disparity with a given finite precision, i.e. disparity estimates are allowed to only take on finite values. Therefore, our estimation of disparity can be modeled as being corrupted with quantization noise with a fixed standard deviation. This constant noise level applies to all disparity values irrespective of the true disparity or correspondingly the actual distance of objects. Let the true disparity be $D_{0}$ and let the observation model for estimated disparity be $D = D_{0} + n$ where $n$ is an additive noise with fixed standard deviation. It will be noted that the dominant component of $n$ is quantization noises. Therefore we have the following relationship
\begin{eqnarray}
Z &=& \frac{fB}{D} = \frac{fB}{D_{0} + n} = \frac{fB}{D_{0}(1 + \frac{n}{D_{0}})} \nonumber \\
\Rightarrow Z &\approx& \frac{fB}{D_{0}}(1-\frac{n}{D_{0}}) = Z_{0} +\frac{-Z^2_{0}}{fB}n \label{Eqn:DepthNoiseModel}
\end{eqnarray}
\begin{figure}
\centering
\subfigure[Scan of plane]{\includegraphics[width=.3\textwidth]{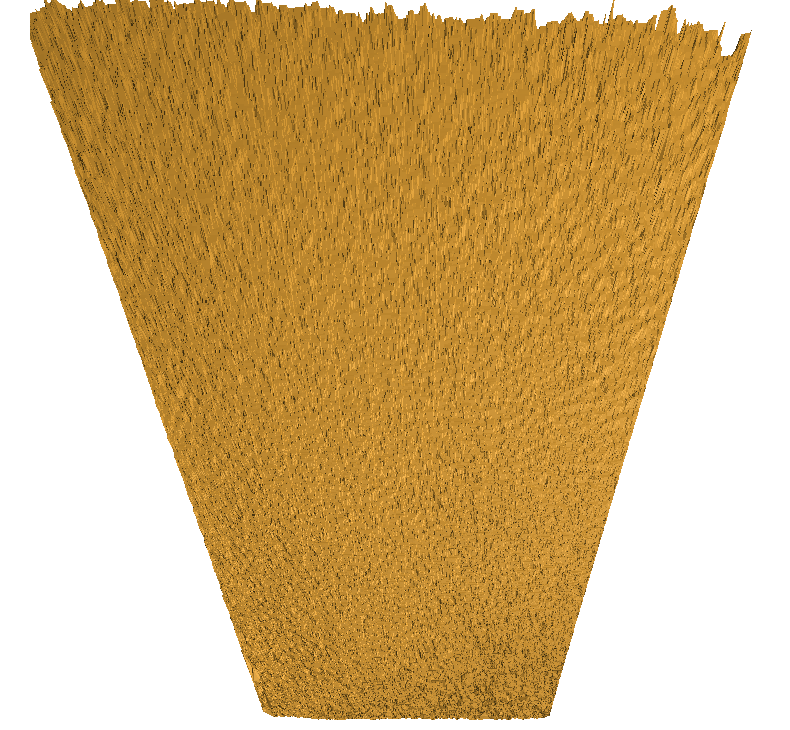}} 
\subfigure[Zoomed view]{\includegraphics[width=.33\textwidth]{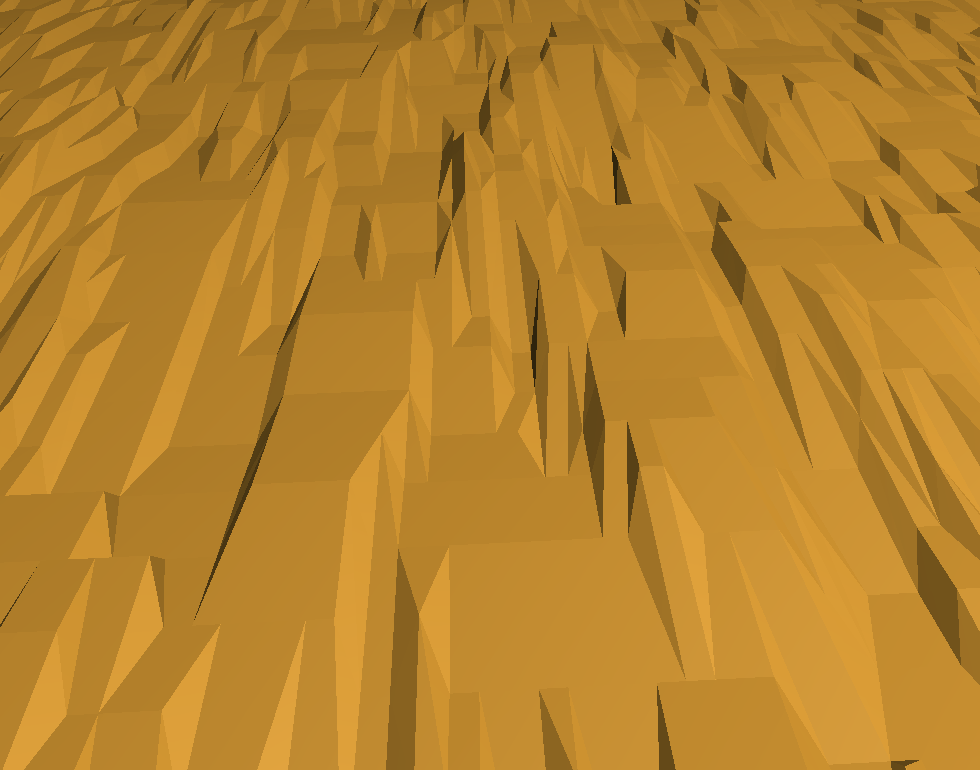}} 
\caption{Scan of a floor plane with Kinect and a zoomed in view that shows the quantization noise in it.}
\label{Figure:FloorPlane}
\end{figure}
\begin{figure*}[t]
\centering
\subfigure[Unique values of depth in plane]{\includegraphics[width=.51\textwidth]{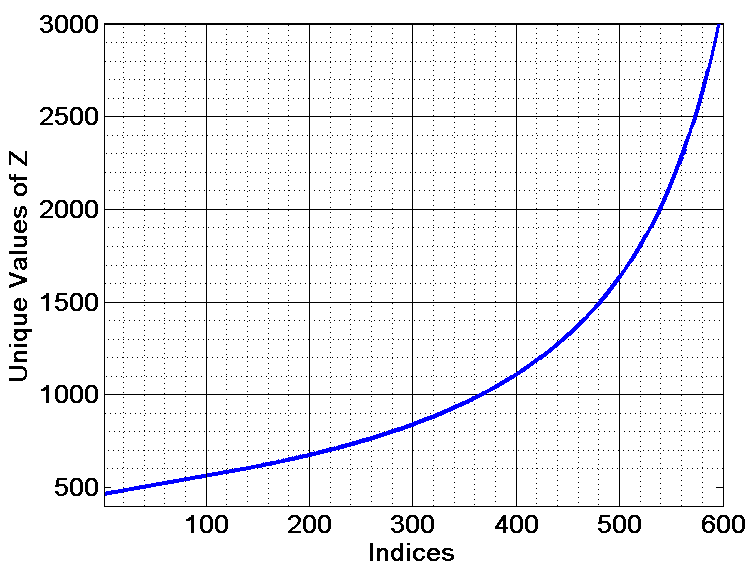}} \\
\subfigure[log-log plot of resolution vs. depth]{\includegraphics[width=.48\textwidth]{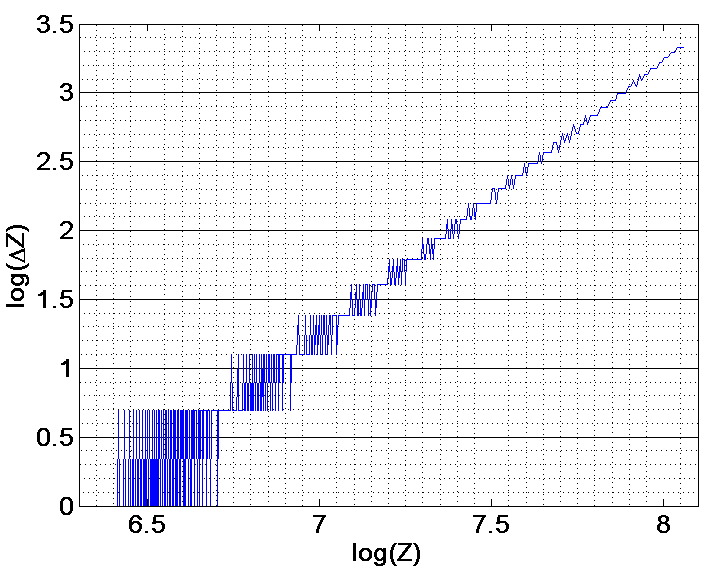}} 
\subfigure[Simulated log-log plot of resolution vs. depth]{\includegraphics[width=.48\textwidth]{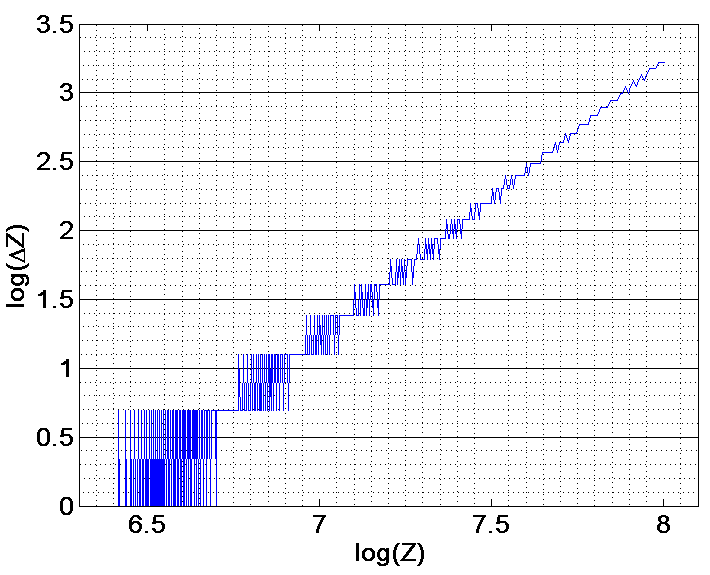}} 
\caption{Analysis of depth resolution of a structured-light stereo range scanner (Kinect). Please see text for details}
\label{Figure:ZResolution}
\end{figure*}
Thus the standard deviation of noise in depth measurement is proportional to the square of the depth of objects. This implies that the precision of estimating depth using a stereo projector-camera pair falls off according to an inverse square law. Equation \ref{Eqn:DepthNoiseModel} can be equivalently derived by differentiating disparity with respect to depth as follows,
\begin{equation}
\frac{\partial D}{\partial Z} = - \frac{fB}{Z^2} \,\,\,\,\Rightarrow\,\,\,\,\, \frac{\partial Z}{\partial D} = -\frac{Z^2}{fB} \label{Eqn:DepthSensitivity}
\end{equation}
i.e. the noise or uncertainty in disparity estimation is amplified by a factor proportional to the square of depth when we convert disparities into depth estimates. To see the implications of this quadratic relationship we can consider the following comparison. The baseline distance between the projector and camera in the Kinect is $B = 75$ mm \cite{KinectROSSpec}. The focal length of the IR camera is found to be 587 pixels in our calibration. This focal length translates to a field-of-view of about $57^\circ$ which is consistent with the available manufacturer specifications \cite{KinectSpec,KinectROSSpec}. Now, if we consider two depth values of say $600$ mm and $1500$ mm (i.e. $2$ feet and $5$ feet approximately) we have the depth sensitivity values of
\begin{eqnarray}
\frac{\partial Z}{\partial D}\bigg\lvert_{Z=600 \textrm{mm}} &=& - \frac{600^{2}}{75 \times 587} = -8.2\, \textrm{mm/pixel} \nonumber \\
\frac{\partial Z}{\partial D}\bigg\lvert_{Z=1500 \textrm{mm}} &=& - \frac{1500^{2}}{75 \times 587} = -51.1\, \textrm{mm/pixel} \nonumber
\end{eqnarray}
i.e., an error of 1 pixel in disparity estimation translates to a depth error of $8.2$ mm and $51.1$ mm at $2$ and $5$ feet respectively.

In Figure \ref{Figure:FloorPlane}(a) we show a 3D scan of a floor obtained using a Kinect. In Figure \ref{Figure:FloorPlane}(b) we show a zoomed-in view of a small region of this scan. As is evident, the Kinect scan is extremely noisy due to the quantization effect in disparity estimation. 
\subsection{Empirical Validation}\label{Subsection:EmpericalValidation}
We now provide empirical validation of this theoretical observation. 
To understand the behavior of such a depth estimate, we extract all unique depth values from the scan of the floor in Figure~\ref{Figure:FloorPlane}. In Figure \ref{Figure:ZResolution}(a) we plot these unique values in ascending order. We may note that for our purposes, we could have carried out this experiment using any object surface as long as the observed object contains a continuous range of depth values. A 3D plane such as a floor is eminently suitable for this purpose. The rate of change between the unique values of $Z$ provides us information on the resolution of the depth estimate.  Specifically, we define the quantity $\Delta Z$ as the difference between adjacent values in the sorted sequence of unique depth values, i.e. $\{\Delta Z\}_{k} = \{Z\}_{k} - \{Z\}_{k-1}$, where $\{Z\}_{k}$ is the $k$-th term in the sorted sequence of unique depth values. If the resolution of the depth estimate was uniform for all $Z$, then the plot of unique values of $Z$ should have been a straight line, i.e. the plot $\{Z\}_{k}$ vs. $k$ should have a constant slope or ${\Delta Z}_{k}$ would have been a constant. However, from Figure \ref{Figure:ZResolution}(a) it is clear that as depth ($Z$) increases the resolution of $Z$ becomes poorer. To understand the nature of the changing depth resolution, in Figure \ref{Figure:ZResolution}(b), we plot the function $\log(\Delta Z)$ vs. $\log(Z)$. The straight line fitted through the observed points has an estimated slope of $1.967$, i.e. a slope almost equal to $2$, thereby empirically verifying our theoretical argument in this paper that the depth sensitivity is proportional to $Z^2$. In other words, for the same perturbation error in disparity measurement, we get a much larger perturbation of depth estimate when $Z$ is larger. We parenthetically remark here that the block-like ringing of $\log(\Delta Z)$ is due to the finite-precision of the depth estimation algorithm implemented on Kinect. Recall that in any disparity measurement in a structured-light stereo configuration, we first estimate the best disparity $D$. This solves for the disparity $D$ with finite resolution, i.e. the estimated $D$ can only take on a discrete set of values specified in units of pixels.

Before proceeding further we need to establish the disparity resolution (in pixels) for the Kinect. It will be noted that the stereo method for measuring disparity would evaluate the stereo matching cost function at a finite number of subpixel disparities. To establish this disparity resolution for Kinect, we conducted a simple experiment. From the Kinect depth map values of the floor plane shown in Figure~\ref{Figure:FloorPlane}(a), we estimate the corresponding disparity $(D=\frac{fB}{Z})$ at every pixel. Subsequently we sort these disparity values in ascending order. In this sorted list, we notice that there are atmost $8$ disparity estimates between two integer disparities, i.e. the subpixel resolution of disparity estimation in the Kinect is $\frac{1}{8}$ pixel. When the distance is large, i.e. disparity is small, there are \textit{exactly} $8$ disparity values between two subsequent integers. When the disparity is high, there are less than $8$ disparity values since the output depth is also quantized to a precision of $1$ mm. Hence, for large disparity (or small depth) two or more consecutive unique disparity values are rounded-off to the same depth estimate due to quantization.

Naturally, when we estimate $Z=\frac{fB}{D}$, we can only get a finite number of depth estimates, i.e. estimated $Z$ cannot be continuous valued. It will also be noted that as $Z$ increases, this `ringing effect' seems to be diminishing as a consequence of the non-linear nature of the log-log plot. To better understand this phenomenon we perform the following synthetic experiment that seeks to replicate the observations depicted in Figure \ref{Figure:ZResolution}(b). We take a sequence of depth values from 50 cm to 3 m with a resolution of 1 mm. We then estimate disparities for these depth values with the known baseline distance and focal length of the Kinect, i.e. $f=587$ pixels and $B=75mm$. Estimated disparities are quantized at $1/8$ pixel resolution as we have established above that the Kinect's disparity resolution is $\frac{1}{8}$ pixel. Then the depths are estimated from these quantized disparities. Subsequently, depth values are also quantized with a precision of $1$ mm. Finally, unique values of depths are extracted and $\log(\Delta Z) $ vs. $\log(Z)$ is plotted in Figure~\ref{Figure:ZResolution}(c) and the corresponding Matlab code is given in Algorithm~\ref{Algorithm:SyntheticResolution}. The synthetic plot in Figure~\ref{Figure:ZResolution}(c) very closely resembles the empirical observation of Figure~\ref{Figure:ZResolution}(b) thereby validating our model for noise in depth estimation using structured-light stereo scanners like the Kinect.
\begin{algorithm}
\caption{Matlab code for simulation of noise-resolution relationship in Kinect depth data. Figure \ref{Figure:ZResolution}(e) was generated using this code. See text for details.} \label{Algorithm:SyntheticResolution}
\begin{algorithmic}[l]
\State 1. b=75;  f=587;
\State 2. z=(b*f./(round(b*f./(500:3000)*8)/8));
\State 3. z=unique(round(z));
\State 4. plot(log(z(1:end-1)),log(diff(z)))
\end{algorithmic}
\end{algorithm}
\section{Applications}\label{Section:Applications}
In Section~\ref{Section:Noisecharacteristic} we have established the fact that the standard deviation of noise in structured-light stereo based depth maps increases proportionally with the squared distance of an object. In this Section we demonstrate the utilization of this fact in  three different applications, i.e. depth map denoising, weighted Volumetric Scan Merging using pixel-wise uncertainty and Plane extraction using the disparity map. We have chosen these three applications as they are amongst the most common approaches of using Kinect depth maps for 3D scene representation and understanding. 
\subsection{Smoothing of Depth Images}\label{Section:AdaptiveSmoothing}
The enhancement of depth maps is an important problem by itself. 
Recent works in depth map enhancement include \cite{essmaeel2014comparative,ChenLL12,cho2013depth,LiuGL12,ZhaoTFTCC13,LoWH13,matyunin2011temporal,WangZPQ14,KimCKA11,milani2012joint,camplani2013depth,qi2013structure}. \cite{milani2012joint,camplani2013depth,qi2013structure,WangZPQ14,LiuGL12,LoWH13,KimCKA11,garcia2013real,ChenLL12} use a guidance color image for smoothing the depth map using the consistency between a depth map and an aligned color image. The use of a guidance image enables these methods to perform hole-filling near the object boundaries. Since the focus of our paper is the noise characteristic of depth maps, we do not use any additional information apart from the depth map itself. Therefore, we do not perform any hole-filling in this paper. In \cite{matyunin2011temporal}, temporal fluctuations of the depth map is used for enhancing the depth estimate. In this paper we emphasize on using the information available in a single depth map to denoise it. In \cite{essmaeel2014comparative} different methods of filtering a depth map are compared. \cite{ZhaoTFTCC13} discusses the application of depth map enhancement for 3D tele-communication. Apart from aesthetic reasons, depth map denoising is a crucial prerequisite for scan registration and other downstream processing in 3D reconstruction. Take for instance, the well-known Iterative Closest Point algorithm which relies on point-to-plane \cite{Masuda02} distance measurements to register 3D scans in a common frame of reference. Such an ICP approach relies on estimating the normal vector at a point on a 3D surface. Normal vectors computed from a scan obtained from a raw depth map are extremely noisy and unreliable. This is so because normal estimation involves discrete differentiation operations which amplifies the noise present in the depth maps. Therefore it is essential to smooth the scan representation for ICP to work effectively. Since it is generally cumbersome to smooth 3D surface representations, such smoothing can be equivalently carried out directly on the depth maps. Indeed, methods such as~\cite{KinectFusion} rely on smoothing the observed depth maps in a preprocessing step.

As in image denoising, depth maps can also be efficiently denoised or smoothed using a neighborhood kernel. Such a local smoothing operation relies on the fact that nearby pixels should have similar intensity (or equivalently depth) values. While typically filters weight the influence of pixels according to their distance from the central pixel, e.g. a Gaussian smoothing filters, we often need to account for intensity or depth discontinuities, i.e. the violation of the assumption that neighboring pixels have similar values. A commonly used modification is the bilateral filter~\cite{BilateralFilter} which modifies the weighting to account for variation of  intensity, thereby effectively carrying out a robust smoothing operation. The standard bilateral filter applied to depth images implicitly assumes that depth values have uniform uncertainty. In the following subsection, we demonstrate how we can incorporate our understanding of depth map noise into the bilateral filter to significantly improve its effectiveness in depth map denoising. This approach is called Adaptive Bilateral Filtering.
\subsection{Adaptive Bilateral Filtering}\label{Subsection:ABL}
Consider an observed depth map $Z(\mathbf{p})$ where $\bfp = (x,y)$ denotes the location of a pixel. The standard approach of bilateral filtering~\cite{BilateralFilter} gives us a denoised depth estimate $\widehat{Z}(\bfp)$ as
\begin{equation}
\widehat{Z}(\bfp)=\frac{1}{W} \sum_{\bfq \in \mathcal{N}(\bfp)} w_{s}(\bfq - \bfp) w_{d}(Z(\bfq)-Z(\bfp)) Z(\bfq) \label{Eqn:BilateralFilter}
\end{equation}
where $w_{s}$ and $w_{d}$ are Gaussian functions for spatial and range weighting with standard deviations of $\sigma_s$ and $\sigma_d$ respectively, $\mathcal{N}(\bfp)$ is the neighborhood of $\bfp$ and $W$ is an overall normalizing factor to have a total sum of $1$ over $\mathcal{N}(\bfp)$. In other words, 
\begin{equation}
w_{s}(\bfx) \propto e^{-\frac{{||\bfx||}^2}{2 {\sigma_s}^{2}} }
\end{equation}
\begin{equation} 
w_{d}(y) \propto e^{-\frac{y^{2}}{2 {\sigma_d}^{2}} }
\end{equation}
In Equation~\ref{Eqn:BilateralFilter}, along with spatial smoothing, the range weight $w_d$ explicitly accounts for the depth differences between the central pixel and other pixels in the support of the smoothing kernel. As a result, using a bilateral filter effectively reduces the influence of neighbors of $\bfp$ that have greatly different values, i.e. violate the implicit assumption that pixels within the support of the smoothing kernel have similar values. Consequently bilateral filtered depth map is smoothened but also preserves depth edges. Although the bilateral filter is an effective strategy, it is inadequate to deal with the problem of smoothing depth maps from structured-light stereo cameras. Recall that the choice of a fixed $\sigma_d$ in the bilateral filter means that we use a fixed `soft threshold' to distinguish between similar neighboring pixels and depth discontinuities. However, as argued in Section~\ref{Section:Noisecharacteristic}, $\sigma_d$ for depth maps is not fixed but varies quadratically with depth. Moreover, since the projector bundle of rays are divergent, the surface sampling density is lower for objects that are far from the scanner. From these two observations, its obvious that for distant objects, the uncertainty of their depth estimate is itself large. In such a scenario, using a fixed $\sigma_d$ is undesirable. For instance, a small $\sigma_d$ means that distant objects are not adequately denoised. Conversely, a large $\sigma_d$ leads to oversmoothing of surfaces that are closer to the depth sensor.
\begin{figure*}
\centering
\mbox{
\subfigure[Raw Scan]{\includegraphics[width=.078\textwidth]{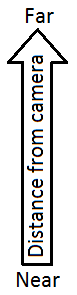}\includegraphics[width=.30\textwidth]{rawfloor.png}}
\subfigure[Bilateral Filtered]{\includegraphics[width=.30\textwidth]{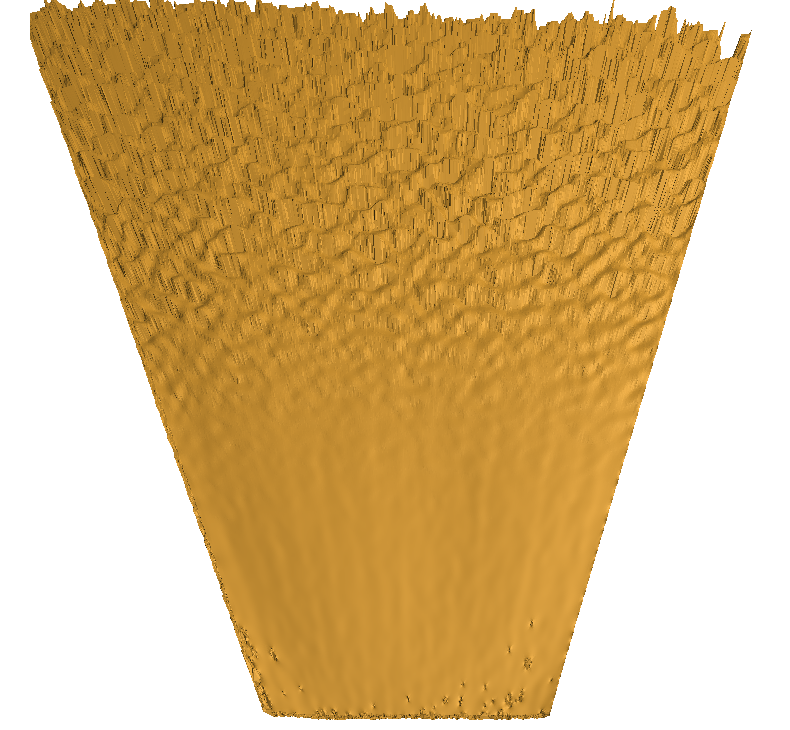}}
\subfigure[Adaptive Bilateral Filtered]{\includegraphics[width=.30\textwidth]{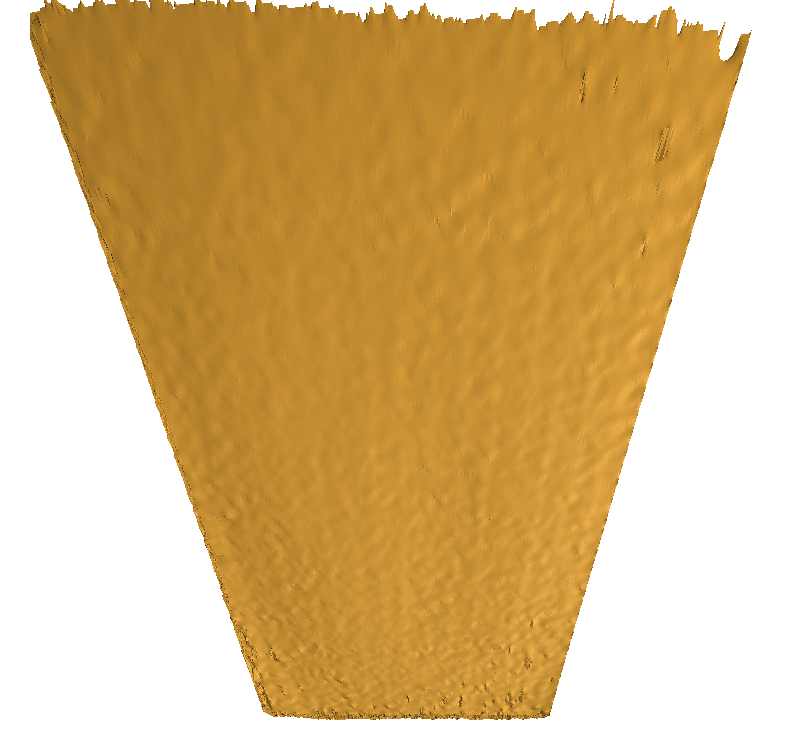}} 
}
\caption{(a) shows the noisy raw scan of a planar surface (corridor floor) (b) shows the result of applying the standard bilateral filter to this scan. Notice that while the lower region, which is closer to the scanner is smoothed, the upper portion which is far away is not adequately smoothed. (c) This problem is mitigated by the use of our adaptive bilateral filter.}
\label{Figure:PlaneFilter}
\end{figure*}
\begin{figure*}
\centering
\subfigure[Raw Scan]{\includegraphics[width=.45\textwidth]{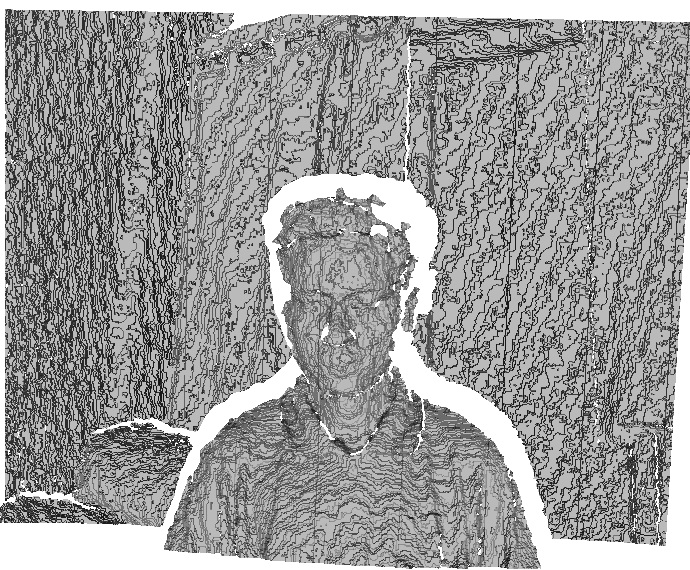}}
\subfigure[Gaussian Filtered]{\includegraphics[width=.45\textwidth]{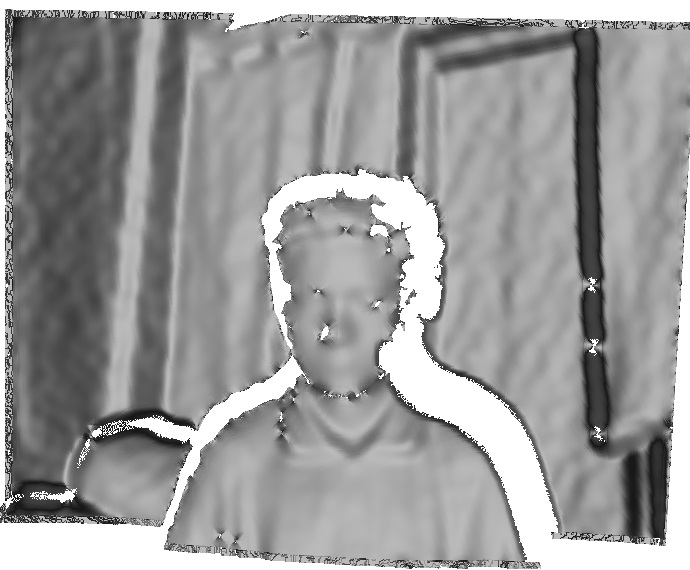}}
\subfigure[Bilateral Filtered]{\includegraphics[width=.45\textwidth]{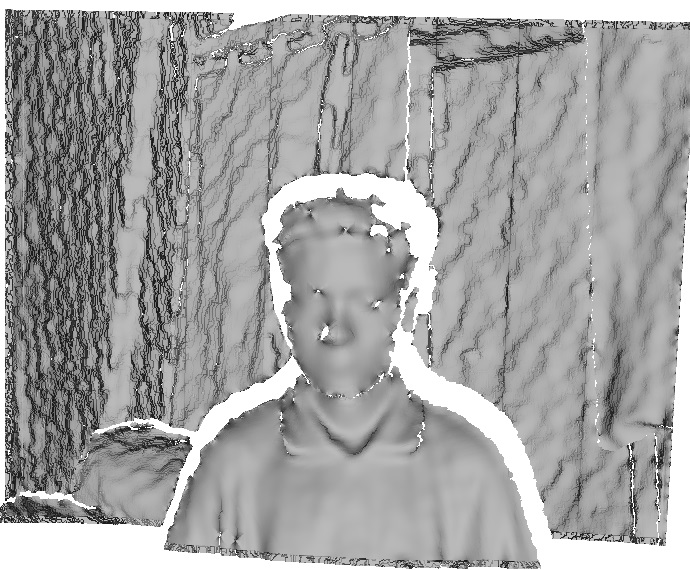}}
\subfigure[Adaptive Bilateral Filtered]{\includegraphics[width=.45\textwidth]{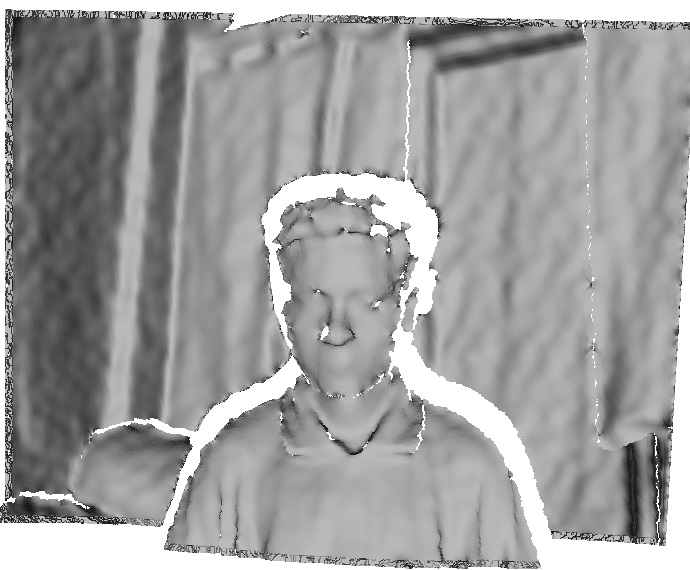} }
\subfigure[Comparison of performance of different filters]{\includegraphics[width=.75\textwidth]{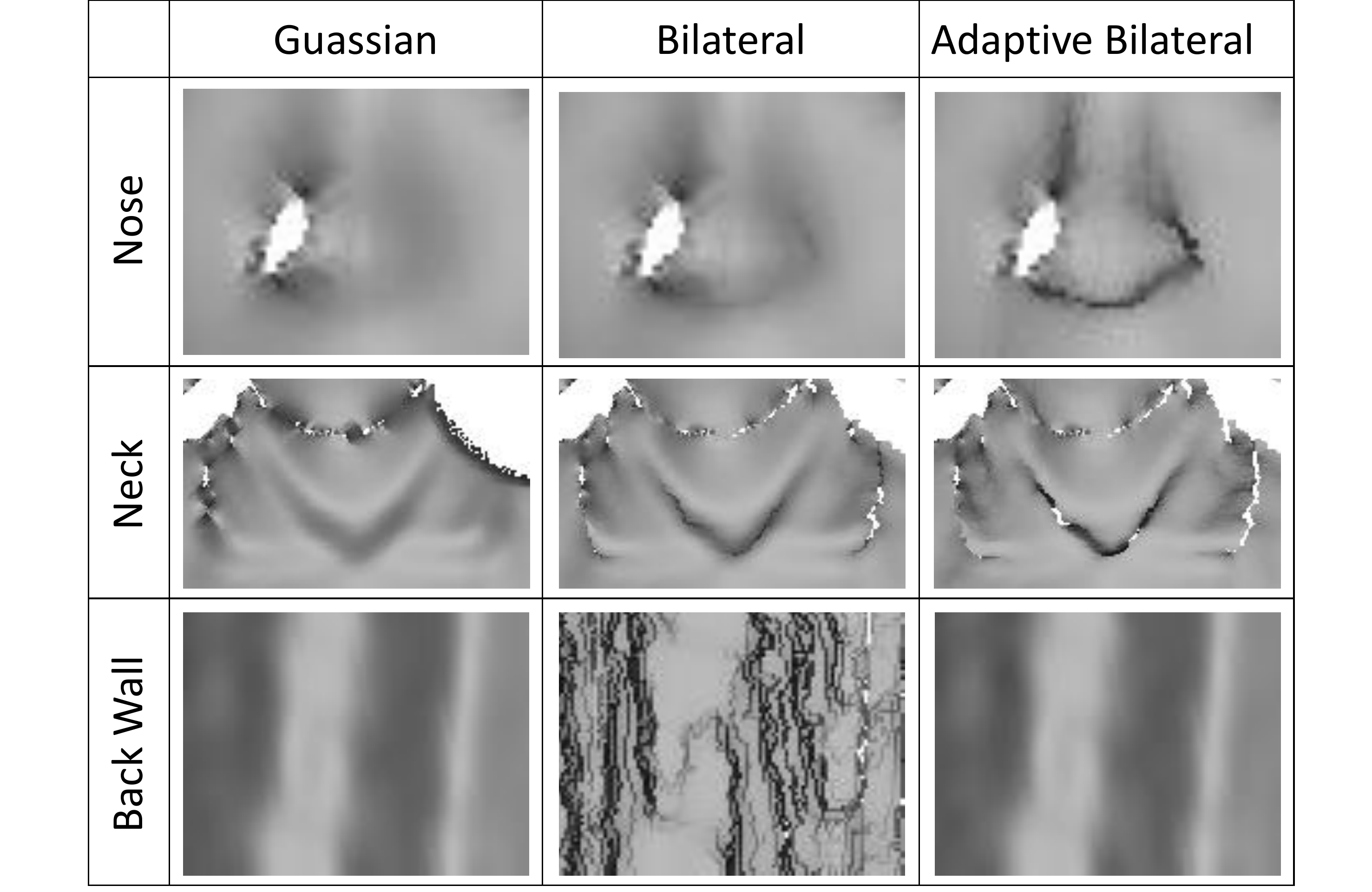}}
\caption{(a) original noisy raw scan of a scene (b) result standard Gaussian filter (c) result of standard bilateral filter (d) result of our Adaptive Bilateral Filter. (e) comparison of the filtering performance of different methods for three zoomed-in regions.}
\label{Figure:BilateralFilterExample}
\end{figure*}
In Figure \ref{Figure:PlaneFilter}(a) we show the mesh representation of a depth image of a planar surface, i.e. the floor of a corridor in our building. We have rotated the mesh representation into an upright view such that the closer part of the floor is at the bottom and the distant part of the floor is at the top of Figure \ref{Figure:PlaneFilter}(a), i.e. depth increase from bottom to top. In Figure \ref{Figure:PlaneFilter}(b), we show the result of applying the standard bilateral filter of equation \ref{Eqn:BilateralFilter} to it. While for a given $\sigma_d$ the standard bilateral filter in Figure \ref{Figure:PlaneFilter}(b) works well for points on the floor that are close to the scanner, we can see that for points that are further away (i.e. the upper part of Figure \ref{Figure:PlaneFilter}(a)) smoothing by the standard bilateral filter is inadequate. Since our analysis shows that the depth estimate sensitivity is quadratically proportional to the depth itself, we modify $\sigma_d$ to vary as $Z^{2}$, i.e. $\sigma_d = k Z^{2}$, where $k$ is a constant. As a result, our modified \textit{Adaptive Bilateral filter} is the same as that of equation \ref{Eqn:BilateralFilter} with the modification that instead of $\sigma_d$ being a constant, for each pixel $\bfp$, we have 
\begin{equation}
\sigma_d(\bfp) = k Z^{2}(\bfp) 
\end{equation}
In Figure \ref{Figure:PlaneFilter}(c) we show the result of applying our adaptive bilateral filter to smoothen the raw depth map shown in Figure \ref{Figure:PlaneFilter}(a). In contrast to the standard bilateral filter, our adaptive filtering shown in Figure \ref{Figure:PlaneFilter}(c) gives superior results for all depths since the smoothing in the range kernel $w_d$ takes into account the specific generative model for the depth image.

We provide an additional example of the superiority of our adaptive bilateral filter in Figure \ref{Figure:BilateralFilterExample} on a more natural depth map of an individual in the foreground with a background consisting of a door and two walls at $90^{\circ}$ to each other. The original raw depth map is shown in Figure \ref{Figure:BilateralFilterExample}(a). Apart from the depth discontinuities between the foreground figure and the background wall, there are also some small holes in the depth map due to occlusions. The result of Gaussian smoothing as shown in Figure~\ref{Figure:BilateralFilterExample}(b) is obtained by choosing a large enough standard deviation to effectively remove the noise in the scan. However, as can be seen from the details shown in the zoomed-in regions in Figure \ref{Figure:BilateralFilterExample}(e) this also results in the blurring of sharp depth edges.

In contrast, the bilateral filter can preserve depth discontinuities (Figure \ref{Figure:BilateralFilterExample}(c)). However, since there are greatly varying depth values in this scene, it cannot provide the appropriate level of smoothing for all parts of the scene. For instance, we can see in Figure \ref{Figure:BilateralFilterExample}(e) that the level of smoothing chosen is inadequate to denoise the back wall which has a higher noise variance. If we had chosen a higher value of $\sigma_d$ here, we would end up oversmoothing the depth map of the individual in the foreground. In contrast, our Adaptive Bilateral Filter automatically tunes $\sigma_d$ to perform the desired smoothing for any depth value. The result of applying our adaptive bilateral filter to the depth image is shown in Figure \ref{Figure:BilateralFilterExample}(d). We can see that both the foreground and background regions are appropriately smoothed while preserving depth discontinuity features since our filter adaptively modifies the standard deviation of the range dimension of the bilateral filter. This observation can also be noted when we look at the different zoomed-in regions shown in Figure \ref{Figure:BilateralFilterExample}(e).
\subsection{Volumetric Scan Merging}\label{Section:ScanMerging}
Scan merging is an important component of any 3D object or scene reconstruction pipeline. For object reconstruction, an object is scanned from different directions. Each scan can view only a part of the object. These scans are then registered or brought into a single frame of reference. In practice, such registration is carried out by utilizing the common features in overlapping regions present in multiple scans. As a consequence once the scans are registered, multiple estimates of the overlapping surface regions are available. These multiple estimates need to be converted into a single unified surface representing the reconstructed object or scene.

This process is called scan merging. Before further discussion on scan merging, to make this article self-contained, we shall briefly discuss how depth maps or range scans are converted into 3D surface representations. Depth maps or range scans are sampled 2D representations of 3D surfaces. The value assigned to a pixel at $(x,y)$ in a depth map represents the depth (distance in the direction of principal axis) $Z(x,y)$ of the 3D point which is imaged at $(x,y)$. Let the focal length of the IR camera be $f$ and the principal point be at $(u,v)$. Then, the depth value of $Z(x,y)$ at the pixel $(x,y)$ represents the 3D point $\bfP = {(X,Y,Z)}^{T}$ such that,
\begin{equation}
\left[\begin{array}{c}X\\Y\\Z\end{array}\right] = \left[\begin{array}{c}\frac{(x-u)Z(x,y)}{f}\\ \frac{(y-v)Z(x,y)}{f}\\ Z(x,y) \end{array}\right] \label{Eqn:DepthMapZ}
\end{equation}
Thus, every pixel in a depth map with a valid depth value represents a 3D point which obeys equation \ref{Eqn:DepthMapZ}. The collection of such 3D points is often called a point cloud and is a sampled description of the scanned 3D surface. After converting depth maps into 3D scans, they are aligned or brought into a single coordinate frame of reference. This process is known as registration and is often carried out using the ICP algorithm and its variants~\cite{EfficientICP}. Finally, multiple co-registered scans of a scene need to be converted or merged into a single unified representation. For scan merging, one relatively simple approach is that of zippering \cite{turk1994zippered}. In zippering, first overlapping parts of scans are eroded and then stitched along the common boundaries. In a final consensus step, vertex positions are re-estimated using the original scans. Another method known as volumetric scan merging \cite{CurlessLevoy} is often the method of choice as it is quite effective under practical situations. In the following, we present a brief overview of volumetric scan merging and readers are referred to \cite{CurlessLevoy} for details.

In volumetric scan merging, we consider a 3D volume that contains or encapsulates all the co-registered scans. For implementational purposes, this 3D volume is discretized into voxels centered on grid points. To achieve a seamless fusion of the multiple measurements in overlapping scan regions, each scan is converted into a truncated-signed-distance-function (TSDF) $f_i(X,Y,Z)$ where the index $i$ denotes the $i$-th scan. The TSDF $f_i(X,Y,Z)$ is defined over the encapsulating volume and is computed at the discrete set of 3D grid points $(X,Y,Z)$ uniformly placed within the volume, i.e. at the center of each voxel. The TSDF magnitude $|f_i(X,Y,Z)|$ at any grid point $(X,Y,Z)$ represents the distance of the nearest point on the $i$-th scan along the corresponding line of sight. To limit the influence of distant surfaces, these distance values are clipped beyond a maximum value, i.e. truncated. The sign of TSDF at a point represents whether the point is nearer or farther from the camera compared to the surface.

Let us consider the $i$-th depth map. Suppose the rotation and the translation for the $i$-th position of the scanner is given by $\bfR_i$ ($3 \times 3$ orthonormal real matrix with determinant 1) and $\bfT_i$ ($3 \times 1$ real vector) respectively. Let the calibration matrix of the scanner be $\bfK_i$ ($3 \times  3$ upper triangular matrix with $\bfK_i(3,3)=1$). In our case, we assume that we use the same scanner throughout, i.e. $\bfK_i = \bfK$.\footnote{We parenthetically note here that IR cameras can be calibrated using the standard procedure of estimating homographies \cite{CalibrationToolBox} by observing a checkerboard pattern illuminated with IR radiation from an incandescent light source. Alternatively, the IR projector of the depth camera can be used as an IR source by placing a semi-transparent sheet of paper in front of the projector.} Then the projection matrix for the $i$-th position of the scanner is given by the $3 \times 4$ matrix 
\begin{equation}
\bfM_i=\bfK_i \left[\begin{array}{c|c}\bfR_i & \bfT_i\end{array}\right] 
\end{equation}
For a 3D point $\bfP = (X,Y,Z)$, its image in the $i$-th depth map $Z_i$ is projected onto the pixel $\bfp_i=(x_i,y_i)$, where $\bfp_i$ satisfies the relationship
\begin{equation}
\left[ \begin{array}{c} x_i\\y_i\\1 \end{array} \right] = \bfM_i \left[ \begin{array}{c} X\\Y\\Z\\1 \end{array} \right] \label{Eqn:Projection}
\end{equation}
where in equation \ref{Eqn:Projection} the equality is a projective relationship~\cite{HartleyZisserman}. Therefore, if the depth value at pixel $\bfp_i=(x_i,y_i)$ in the $i$-th scan is $Z_i(x_i,y_i)$, we can identify it with a 3D point with the coordinates (with respect to $i$-th cameras local coordinate system) given by
\begin{equation}
\left[\begin{array}{c}X_i\\Y_i\\Z_i\end{array}\right] = \left[\begin{array}{c}\frac{(x_i-u_i)Z_i(x_i,y_i)}{f_i}\\ \frac{(y_i-v_i)Z_i(x_i,y_i)}{f_i}\\Z_i(x_i,y_i)\end{array}\right]
\end{equation}

Clearly, denoting the 3D point as $\bfP = {(X,Y,Z)}^{T}$, the distance of the surface from the center of the camera in the $i$-th position along the line of sight passing through the pixel $(x_i,y_i)$ is given by ${\left|\!\left|\bfP_i\right|\!\right|}$. Also, the distance of the grid point $\bfP = {(X,Y,Z)}^{T}$ from the center of the camera in $i$-th position is given by ${\left|\!\left|\bfR_i \bfP + \bfT_i\right|\!\right|}$ where $\left|\!\left|.\right|\!\right|$ is the 2-norm. Therefore, the TSDF is given as
\begin{equation}
f_i(X,Y,Z)=\max{ \left(\min{\left(\left|\!\left|{\bfR_i \bfP +\bfT_i}\right|\!\right| - \left|\!\left|\bfP_i\right|\!\right|,f_{max}\right)} , f_{min} \right) }
\end{equation}
where $f_{min}$ and $f_{max}$ are the minimum and maximum values of TSDF's, beyond which TSDF's are clipped. The TSDF's $f_i(X,Y,Z)$ from all the scans are then summed up with appropriate weights $w_i(X,Y,Z)$, leading to a unified representation
\begin{equation}
F(X,Y,Z)=\frac{\sum_i{w_i(X,Y,Z) f_i(X,Y,Z)}}{\sum_i{w_i(X,Y,Z)}} \label{Eqn:SummedTSDF}
\end{equation}

The zero crossing surface of $F(X,Y,Z)$ in the encapsulating volume is the merged and unified representation of all the co-registered scans. This unified representation is obtained by performing an iso-surface extraction which is in practice implemented using a marching cube algorithm~\cite{lorensen1987marching,montani1994modified}. It will be noted that the process of summing up of individual TSDF's effectively averages the position of surfaces with multiple observations. This helps in denoising or reducing the uncertainty of vertex positions. However, to correctly account for the information present in each observation, the different measurement representations $f_i(X,Y,Z)$ should be weighted by an appropriate function $w_i(X,Y,Z)$ that reflects its uncertainty. Consider the scenario where a surface might be observed from two different distances. As noted earlier, the standard deviation of the observation made from a location further from the surface is higher compared to that made from a closer position. Consequently, unless the weighting function accounts for the relative accuracy of these observations, in the superimposed TSDF function $F(X,Y,Z)$ of equation \ref{Eqn:SummedTSDF}, the precision and details present in the closer observation will be lost due to the inordinate influence of the noisy distant observation. Since from Equation~\ref{Eqn:Projection} we can see that the point $(X,Y,Z)$ projects to $(x_i,y_i)$ in scan $i$, the corresponding weight $w_{i}(X,Y,Z)$ is derived from the uncertainty of the depth value at the pixel $(x_i,y_i)$.

Now, let us consider a set of scalar observations $x_i = \mu + n_i$ where the noise terms $n_i$ are Gaussian independently distributed, i.e. $n_i \sim N(0,\sigma^{2}_{i})$ with varying $\sigma_i$. It can be easily seen that the maximum likelihood estimate (MLE) for $\mu$ is given by $\widehat{\mu} \propto  \displaystyle\sum_i \frac{x_i}{\sigma^2_i}$. In other words, each individual measurement $x_i$ is weighted by a factor inversely proportional to the variance of that observation. Since in our scenario of depth map measurements, the scale depth estimate $Z(x_i,y_i)$ has standard deviation proportional to ${Z(x_i,y_i)}^{2}$. Therefore, the weights we assign are inversely proportional to the fourth power of depth, i.e.
\begin{equation}
w_i(X,Y,Z)=w_i(x_i,y_i)=\frac{1}{{Z_i(x_i,y_i)}^4} \label{Eqn:Weight}
\end{equation}

It will be evident from equation \ref{Eqn:SummedTSDF} that any scalar constant factor has no influence on the summed TSDF. As a result, we need not incorporate any normalization constant in equation \ref{Eqn:Weight} and can simply equate $w_i$ with the inverse of the fourth power of depth $Z_i$. While summing up TSDF's at a grid point each TSDF is weighted by the weight assigned to them at that point. Under the assumption of orthographic projection and using the fact that the range errors are independently distributed along the sensors line-of-sight, it can be shown that the volumetric scan merging method is optimal in a least square sense~\cite{CurlessLevoy}. Hence, with these assumptions, our weighting scheme gives a maximal likelihood estimate of the surface when we model the depth estimates to have Gaussian noise with standard deviation quadratically varying with the distance of an object.
\begin{figure*}
\centering
\subfigure[]{\includegraphics[width=.48\textwidth]{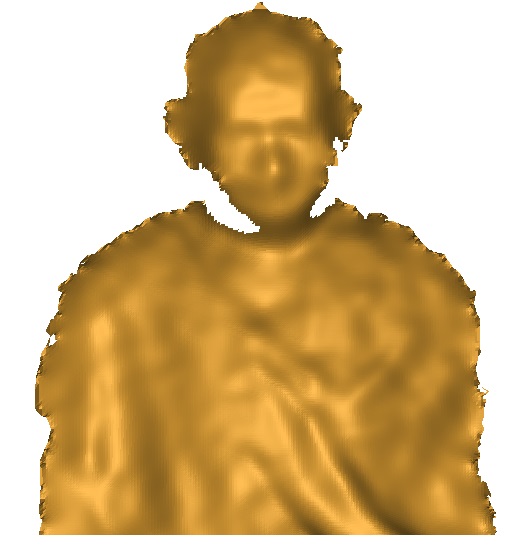}}
\subfigure[]{\includegraphics[width=.48\textwidth]{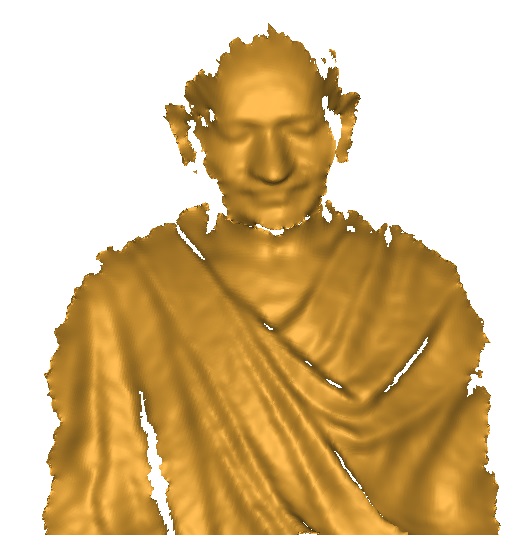}}
\subfigure[]{\includegraphics[width=.48\textwidth]{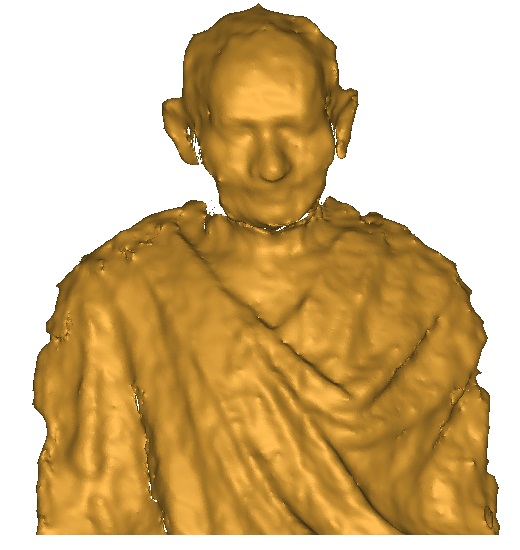}}
\subfigure[]{\includegraphics[width=.48\textwidth]{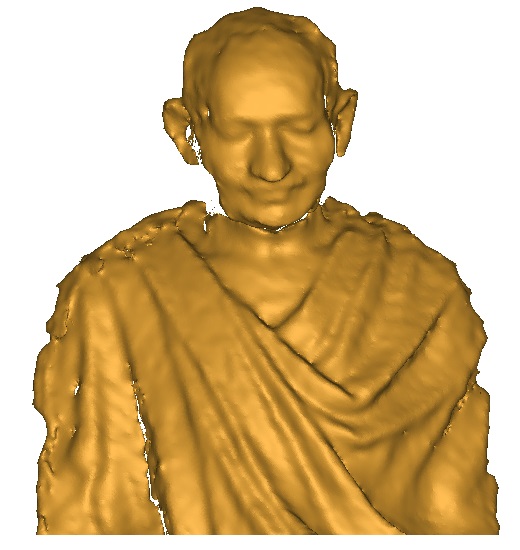}}
\caption{Scan of Mahatma Gandhi's statue taken from long (a) and short (b) distances. Results of merging (a) and (b) with unweighted and our weighted volumetric method are depicted in (c) and (d) respectively.}
\label{FigureWeightedVCG}
\end{figure*}

We demonstrate the effectiveness of the weighting scheme of equation \ref{Eqn:Weight} with the following experiment. We scanned a life-size statue of Mahatma Gandhi located on the grounds of Sabarmati Ashram in Ahmedabad from varying distances. In Figure \ref{FigureWeightedVCG}(a) we show a scan taken from close range, i.e. an average distance of approximately $75$ cm whereas in Figure \ref{FigureWeightedVCG}(b) we show a scan of the same statue taken from a greater distance of about $1.5$ m. While the area of the statue covered by the more distant scan is larger than the nearer one, for simplicity of visualization we crop the depth maps to show the same common parts of the statue. Figure \ref{FigureWeightedVCG}(c) is the result of the standard volumetric merging of these two scans (i.e. with weights $w_i=1$) and Figure \ref{FigureWeightedVCG}(d) is the result of volumetric merging using our weighting scheme. It is evident from a comparison of Figure \ref{FigureWeightedVCG}(c) and Figure \ref{FigureWeightedVCG}(d) that our weighting scheme provides a much better result than standard volumetric merging. This is because the scan in Figure \ref{FigureWeightedVCG}(b) is taken from a closer distance and has more details than the distant scan of Figure \ref{FigureWeightedVCG}(a). If they are averaged with equal weights, then all the details of the closer scan are sacrificed. In contrast, our weighting scheme preserves these details as it naturally assigns greater weightage to the observations taken from the closer scan.
\begin{figure*}
\centering
\subfigure[Unweighted VCG Reconstruction]{\includegraphics[width=.48\textwidth]{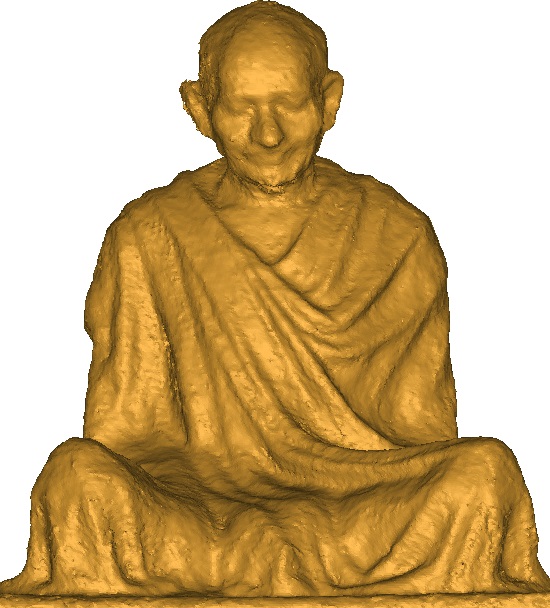}}
\subfigure[Our Weighted VCG Reconstruction]{\includegraphics[width=.48\textwidth]{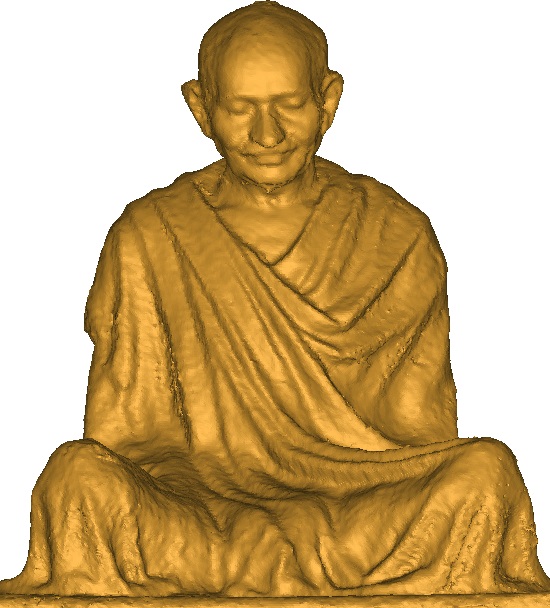}}\\
\subfigure[Different views of our Weighted VCG Reconstruction]{
\includegraphics[width=0.32\textwidth]{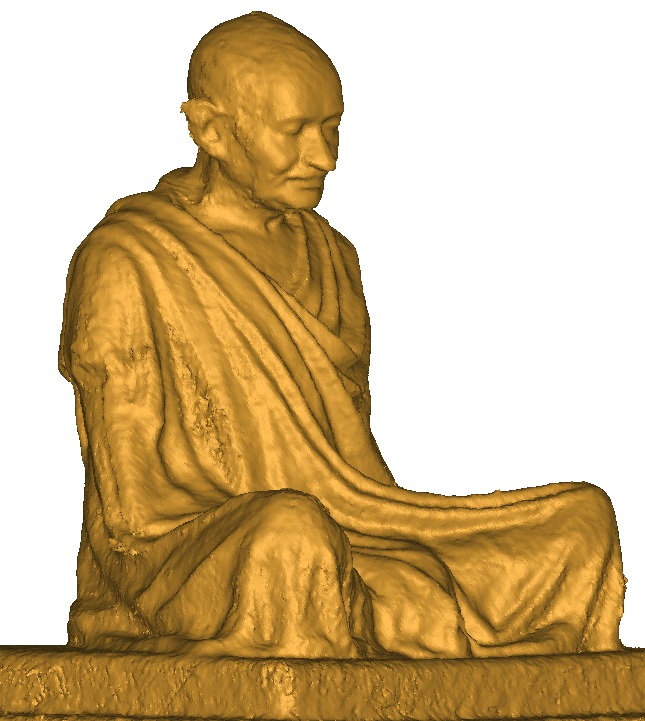}
\includegraphics[width=0.32\textwidth]{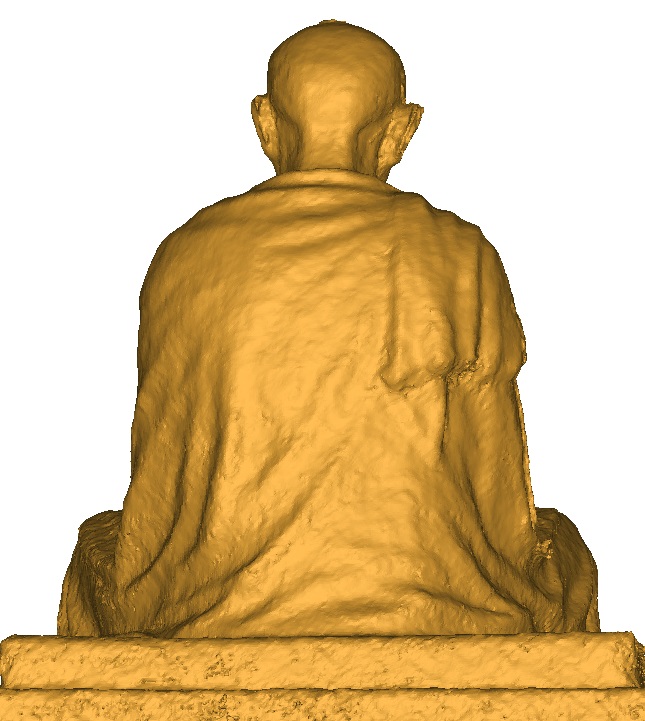}
\includegraphics[width=0.32\textwidth]{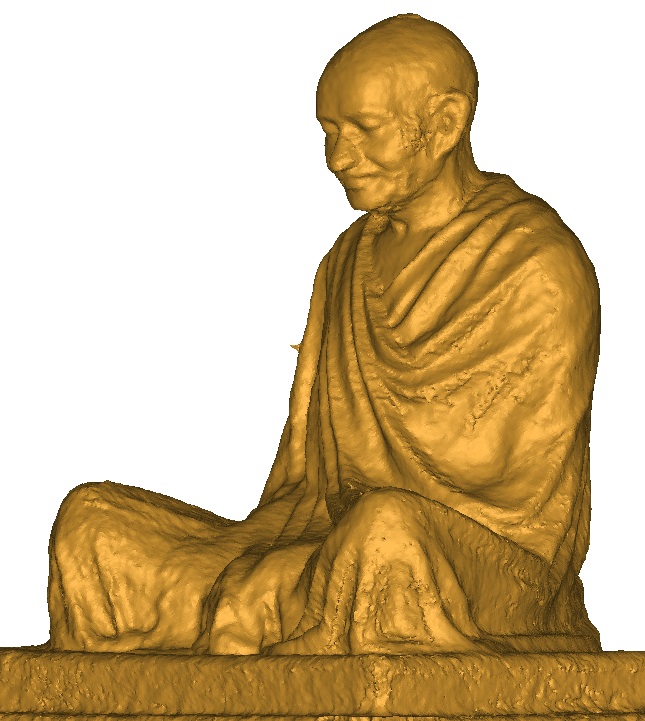}}
\caption{Our reconstruction of a life-size statue of Mahatma Gandhi.} 
\label{Figure:Mahatma}
\end{figure*}

In Figure \ref{Figure:Mahatma} we show the complete 3D reconstruction of the life-size statue of Mahatma Gandhi. This complete 3D model is built with 29 scans taken by walking around the statue. As this statue is located outdoors we acquired the scans at night since commercially available IR depth cameras do not work in the presence of moderate to strong sunlight. As mentioned earlier, this model is scanned from widely varying distances. While the scans taken from a distance covers a larger portion of the statue and hence help in accurate registration, scans taken from a closer position have much better detail. Closer scans have a greater amount of detail since the uncertainty in depth estimates is lower and at the same time a small surface area is imaged over a larger number of depth map pixels (compared to a long range scan), thereby providing a larger number of detailed surface measurements. Therefore, for successful 3D modeling, we need to allow for both types of scans. Nevertheless, as seen earlier, while merging scans taken from different distances, it is essential to weight them according to their inherent accuracy to achieve the best possible 3D reconstruction. As can be seen from Figure \ref{Figure:Mahatma}(b-c), this goal is successfully attained with our weighted volumetric merging scheme. In contrast, we can see from Figure \ref{Figure:Mahatma}(a) that the standard unweighted volumetric scan merging does poorly in comparison as it does not assign a lower level of influence to scans taken from a greater distance.
\begin{figure*}
\centering
\includegraphics[width=.85\textwidth]{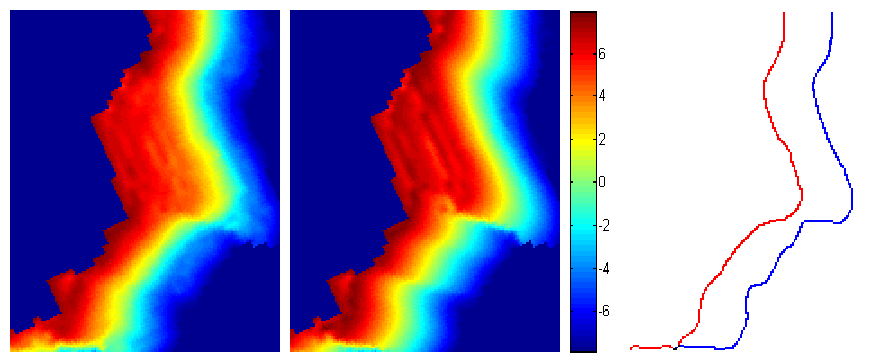}
\caption{Comparison of TSDF's for ordinary volumetric merging (left) with our weighted volumetric merging  (middle). The extracted zero crossings are also shown (right). Zero crossing for unweighted and weighted TSDF's are shown in red and blue respectively.}
\label{FigureWeightedVCG2}
\end{figure*}

In Figure \ref{FigureWeightedVCG2}, we show a cross section of the TSDF functions generated with the ordinary unweighted approach and our weighted averaging schemes for this example. It is clear that our approach produces a function which better preserves details compared to the ordinary averaging approach. For instance, in Figure \ref{FigureWeightedVCG2} the outline of the lips is seen to be more prominent in our approach. This observation has been further illustrated in  Figure \ref{FigureWeightedVCG2}(c) which compares the zero crossing curves extracted from both the TSDF's. Here it is easily seen that our weighting is able to preserve details better than the unweighted volumetric scan merging approach.
\begin{figure*}
\centering
\subfigure[Different views of our 3D model]{
\includegraphics[width=.191\textwidth]{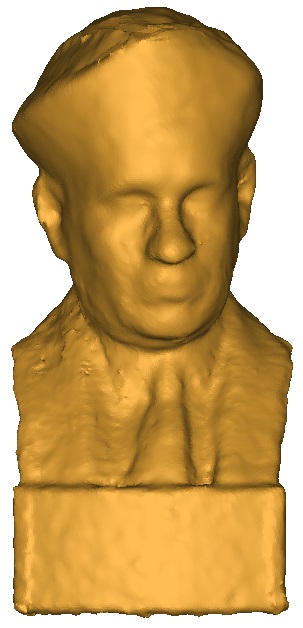}
\includegraphics[width=.2016\textwidth]{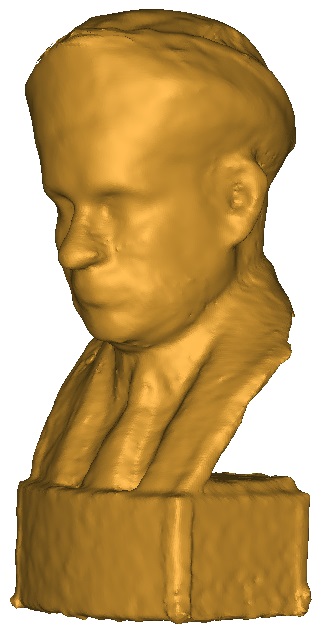}
\includegraphics[width=.186\textwidth]{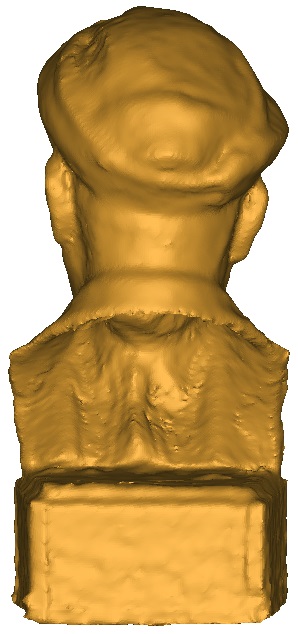}
\includegraphics[width=.1976\textwidth]{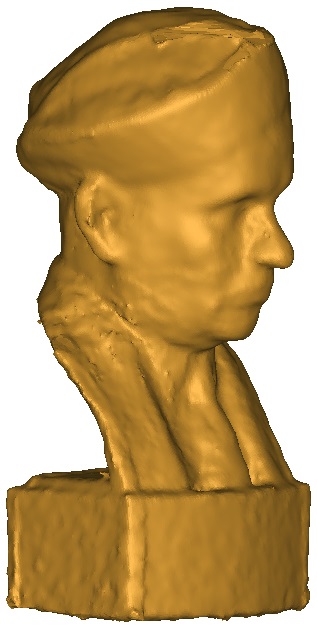}}
\subfigure[Kinect Fusion]{
\includegraphics[width=0.192\textwidth]{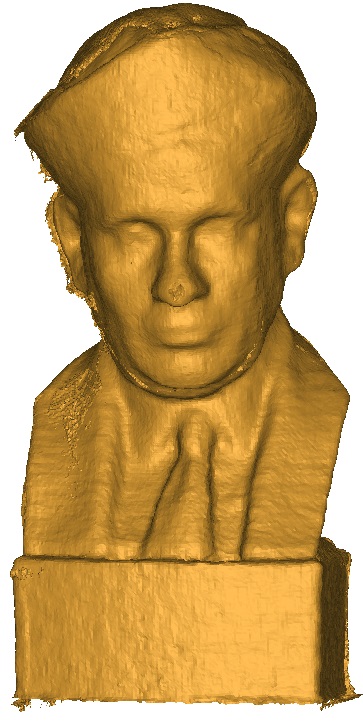}}
\caption{(a) shows the reconstructed 3D model of a bust of the scientist C. V. Raman rendered from different viewpoints; (b) shows the reconstruction obtained using Kinect Fusion.}
\label{Figure:CVRaman}
\end{figure*}

As an another example of the applicability of both our depth map denoising and weighted volumetric scan merging, in Figure~\ref{Figure:CVRaman} we show the results of our reconstruction of a metal bust of height of about $50$ cm of the Indian scientist, C. V. Raman. The reconstructed model is shown from different viewpoints in Figure~\ref{Figure:CVRaman}(a). We first filtered the individual depth maps using the adaptive bilateral filter detailed earlier in Section \ref{Subsection:ABL}. These smoothed depth maps are converted into 3D scans and registered globally in a single frame of reference using the motion averaged ICP (MAICP) algorithm presented in  \cite{VenuPooja2014}. It is worth reiterating here that the adaptive bilateral filtering step is essential to make the subsequent registration steps work as the raw depth maps from the Kinect are very noisy. Without this filtering step, the normals estimated on the scans would be error prone and result in registration failures. Once the registration of all scans is achieved, we use our modified volumetric scan merging approach detailed in Section \ref{Section:ScanMerging} to generate a final 3D model. Figure~\ref{Figure:CVRaman}(b) is the corresponding 3D reconstruction obtained using Kinect Fusion~\cite{KinectSDK}. As can be seen, the quality of 3D reconstruction using our approach is similar compared to Kinect Fusion that needs to capture depth maps at video rate and needs expensive hardware to process them.  Also Kinect Fusion requires the depth camera to be moved very slowly and carefully during the scanning process. In this experiment, Kinect Fusion was run for more than 5 minutes at a frame rate of around 3-4 fps, i.e. more than $1000$ frames were captured and processed in the reconstruction. In contrast, our result is obtained using as few as $21$ depth maps. Yet our result is comparable to that of Kinect Fusion because we take care of the uncertainties in individual measurement in an optimal way.
\subsection{Plane Extraction}\label{Section:PlaneExtraction}
The ease of use of depth cameras like Kinect have also made it suitable for use in mobile robotics applications such as simultaneous localization and mapping (SLAM) \cite{engelhard2011real,EndresHSCB14}, navigation etc. In this subsection, we will focus our attention to a specific subproblem in a typical SLAM pipeline, i.e. of extracting 3D planes from noisy range images. 3D planes are ubiquitous in indoor scenes and can be utilized in a variety of ways for aiding motion estimation and tracking \cite{TaguchiJRF12,DouGFF12,ataer2013tracking}. Some approaches to  plane extraction from noisy range images are presented in \cite{poppinga2008fast,holz2012real,pathak2010uncertainty}. ~\cite{poppinga2008fast} describes a method for plane extraction on the range maps obtained from time-of-flight scanners. This method finds candidate planes by means of an expensive region-growing algorithm and the sequential nature of range videos is exploited to achieve speed up.~\cite{holz2012real} proposes a fast algorithm for plane extraction using RGB-D cameras like Kinect, wherein surface normals are computed using integral image and clustered to find planar regions. But in \cite{holz2012real} the intrinsic noise characteristics of structured-light stereo depth cameras is not exploited.~\cite{pathak2010uncertainty} describes plane extraction from the range map where the noise is a quadratic function of both the distance and the incidence angle.

As in many pattern recognition applications, given a hypothesis of a 3D plane, we can propose a test for whether a given 3D point belongs to this plane. Typically this would involve measuring the distance of the point to the plane and comparing it with a fixed threshold. The threshold is determined by the expected level of noise in the observed 3D point location. However, in the case of depth map observations each 3D point has a different amount of noise associated with it, implying that we cannot use a fixed threshold in our plane fitting hypothesis test. It is evident that points that are at a greater distance need a more relaxed distance threshold in contrast with points that are closer to the depth camera. While such a varying threshold can be incorporated into the test for plane fitting, it will be recognized that we need to hypothesize the parameters of the 3D plane based on the observed data itself. In such a scenario, the non-uniformity of the standard deviation of observation noise makes it hard to generate valid 3D plane hypotheses. However, we can avoid this difficulty if we use disparity maps instead of depth maps.

In Section \ref{Section:Noisecharacteristic} we had derived the fact that the standard deviation of depth measurement varies quadratically with depth. However, this observation model is itself an outcome of the assumption that the uncertainty (i.e. standard deviation) in disparity measurement is a constant and independent of depth or disparity. Therefore, for the purposes of 3D plane extraction it is particularly advantageous to work with disparity maps instead of depth maps. This is true since, for a planar surface, the 2D disparity map $D(x,y)$ has an affine relationship with the pixel location $(x,y)$. Consider a point $\bfP=(X,Y,Z)$ on a 3D plane satisfying the equation $aX+bY+cZ+1=0$. If the point $(X,Y,Z)$ gets projected to pixel location $(x,y)$ in the IR camera, then, 
\begin{equation}
x=\frac{fX}{Z}+u;\,\,\,\,y=\frac{fY}{Z}+v
\end{equation}
where $f$ is the focal length of the camera and $(u,v)$ is the principal point or focal point of the camera. The disparity at the point $(x,y)$ is given by $D(x,y)=\frac{fB}{Z}$, where $B$ is the baseline distance between the projector and the camera center. Substituting variables in the equation of the plane we have,
\begin{eqnarray}
& & aX+bY+cZ+1=0 \nonumber \\
& \Rightarrow & a\frac{fX}{Z}+b\frac{fY}{Z}+cf+\frac{f}{Z}=0 \nonumber \\
& \Rightarrow & a(x-u)+b(y-v)+cf+\frac{D(x,y)}{B}=0 \nonumber \\
& \Rightarrow & ax+by+\frac{1}{B}D(x,y)+(cf-au-bv)=0 \label{Eqn:AffineDepthMap}
\end{eqnarray}

The affine relationship in equation \ref{Eqn:AffineDepthMap} helps in formulating a very efficient algorithm for plane extraction using a disparity map. The disparity image is first passed through a Laplacian-of-Gaussian (LoG) filter which is a smoothened version of the 2D Laplacian operator and has a high response for sharp changes and a zero response for regions that display a linear variation. The LoG filter is popularly used for edge or blob detection. Since the disparity map obeys an affine relationship in planar regions, the ideal response of a LoG filter for a planar region should be equal to zero. Conversely, for non-planar regions we can expect to observe a higher response when a LoG filter is applied. Therefore, in our case, LoG response values above a given threshold are considered to belong to a non planar region. It will be noted here that this threshold is a fixed constant, i.e. independent of depth or equivalently disparity. Amongst all pixels that pass this test, we find connected components to identify planar regions in the depth maps. For each planar region, the parameters of the plane are computed robustly using the relationship in equation \ref{Eqn:AffineDepthMap}. Consequently, the use of a fixed threshold on the disparity map helps us in formulating a very fast technique for plane detection in depth maps. 

Once the parameters for a 3D plane are estimated, we find more points that closely satisfy the relationship in equation~\ref{Eqn:AffineDepthMap}. The parameters of the 3D planes are re-estimated using all the fitted points and planes which have very close values of parameters are merged into composite planes. We repeat this process of estimating parameters and the set of points that fit the plane iteratively akin to that of k-means which is a classic iterative algorithm for clustering of points. The k-means approach requires the specification of the number of clusters. In our plane extraction algorithm, we estimate the number of planes from the number of connected components present after LoG filtering and thresholding. In this context, when plane parameters are close to each other, we merge them into single planes. If a point fits multiple plane hypotheses equally well, we use its neighboring pixels in the depth map to disambiguate, i.e. we try and find the largest coherent regions corresponding to 3D planes in a depth map.
\begin{figure*}
\centering
\subfigure[]{\includegraphics[width=.3\textwidth]{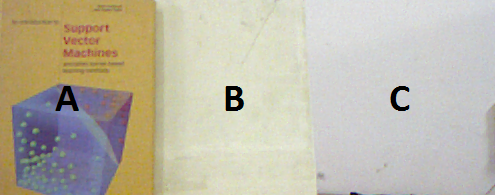}}
\subfigure[]{\includegraphics[width=.3\textwidth]{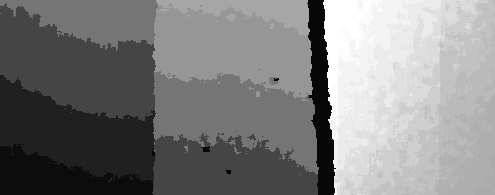}}
\subfigure[]{\includegraphics[width=.3\textwidth]{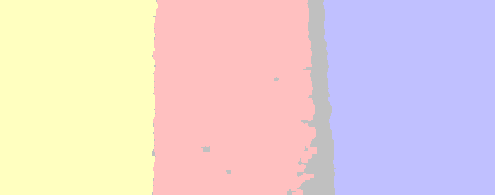}}\\
\subfigure[]{\includegraphics[width=.8\textwidth]{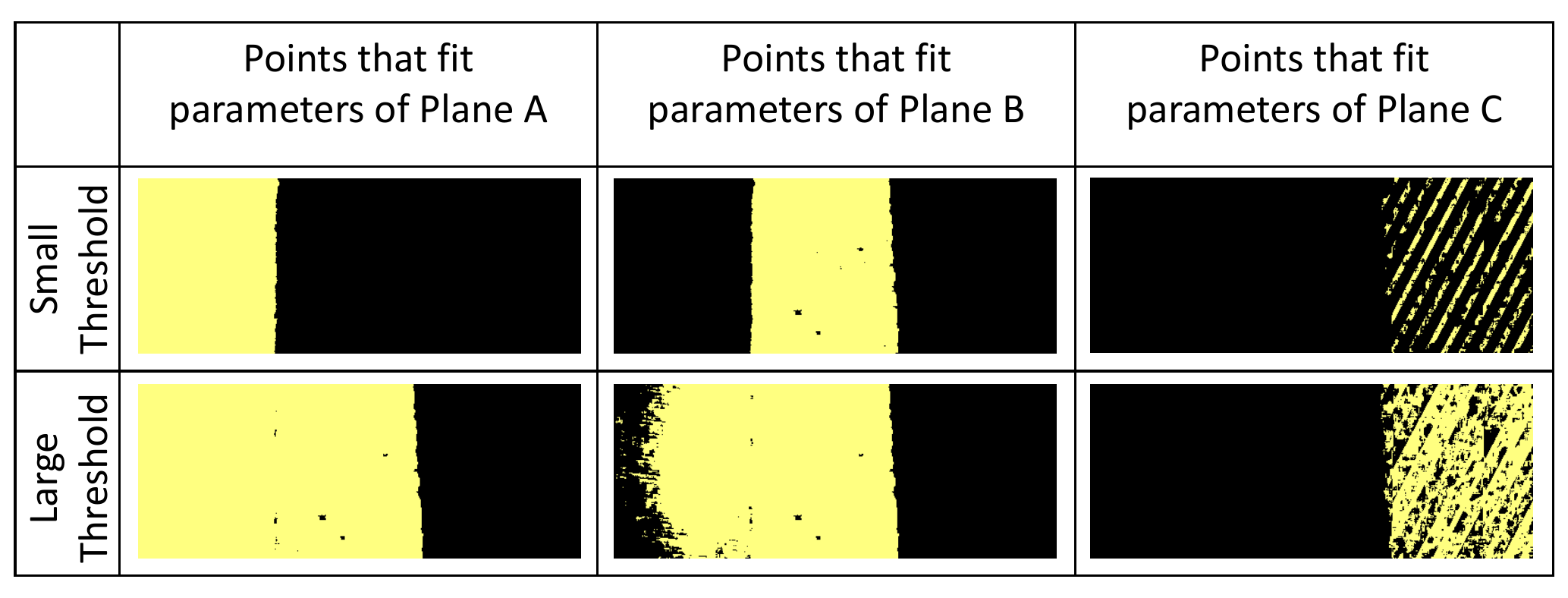}}
\caption{(a) Color image of a scene consisting of three planes. (b) Depth map of the three planes (histogram equalized for ease of visualization). (c) Result of our plane extraction method. Three extracted planes are marked with three colors. Gray regions do not belong to any of the planes. (d) Result of using a fixed threshold based plane fitting. This clearly shows that neither a small or large threshold is adequate for all three planes. Please see the text for details.}
\label{Figure:3Planes}
\end{figure*}

Before we present results of our plane extraction method, we provide a simple experiment that clearly demonstrates the fact that a fixed distance approach to plane fitting is inadequate to handle different situations. In Figure~\ref{Figure:3Planes}(a) and (b) we show an RGB and depth map respectively of a scene consisting of $3$ planes. The labeled planes A, B and C are in increasing order of distance from the depth camera. While plane C is at a distance of about $10$ feet from the depth camera, planes A and B are located much closer at a distance of $2$ feet. The depth map in Figure~\ref{Figure:3Planes}(b) is histogram equalized for ease of visualization. Finally, planes A and B are at about $2$ feet from the depth camera, but differ in depth by $1$ cm, i.e. differ in the thickness of the book shown in plane A. In Figure~\ref{Figure:3Planes}(c) the result of our plane extraction is shown. Clearly, all three planes are correctly and distinctly identified.

In Figure~\ref{Figure:3Planes}(d) we demonstrate why a fixed distance threshold based approach is not adequate. To obtain the best possible parameter estimates for the $3$ planes we manually segment the three planes in the depth map and separately compute the parameters of the three planes using principal component analysis on the corresponding 3D points for each plane. Subsequently, we specify a threshold $\bfT$ which determines whether a 3D point belongs to a given plane. If a 3D point is within distance $\bfT$ from a given plane, we may classify it as belonging to that plane. For our test we used two different values of $\bfT$, i.e. $\bfT=5$ mm and $\bfT=20$ mm. From Figure~\ref{Figure:3Planes}(d) we see that when $\bfT=5$ mm, there is a good fit for most points in planes A and B with their respective plane models. However, most points on plane C do not satisfy this test. Since plane C is further from the camera, the noise in individual depth points is large, hence for a small threshold $\bfT$, most points on plane C would be declared to have failed to fit the model for plane C, i.e. they would fail to be classified as belonging to plane C. However, although many points on C are incorrectly classified as not belonging to C, for the small threshold $\bfT$, no point is wrongly classified as belonging to a different plane, i.e. points are distinguishable since the small threshold $\bfT$ is a stringent test. When we use the larger threshold $\bfT=20$ mm, we see in Figure~\ref{Figure:3Planes}(d) that we are no longer able to correctly classify points into planes A and B. In fact, since the classification threshold $\bfT$ is relaxed (i.e. larger), we can no longer distinguish plane A from B. However, in this case, most of the points on plane C are correctly classified as they now lie within the larger distance threshold of $\bfT=20$ mm. Hence, from this experiment we may conclude that we cannot use either a small or a large fixed threshold to correctly classify points on planes under different scenarios. In contrast, since we take into account the true uncertainty of depth measurements, our approach can be seen to correctly classify the points into three distinct planes in Figure~\ref{Figure:3Planes}(c). We now proceed to demonstrate the efficacy of our method with more results in the next subsection.
\subsection{Plane Extraction Results}
\begin{figure*}
\centering
\subfigure[Plane extraction on two laboratory scenes. Left to right: Depth map, RGB image and the result of plane extraction. Black regions are identified as being non planar.]{\includegraphics[width=\textwidth]{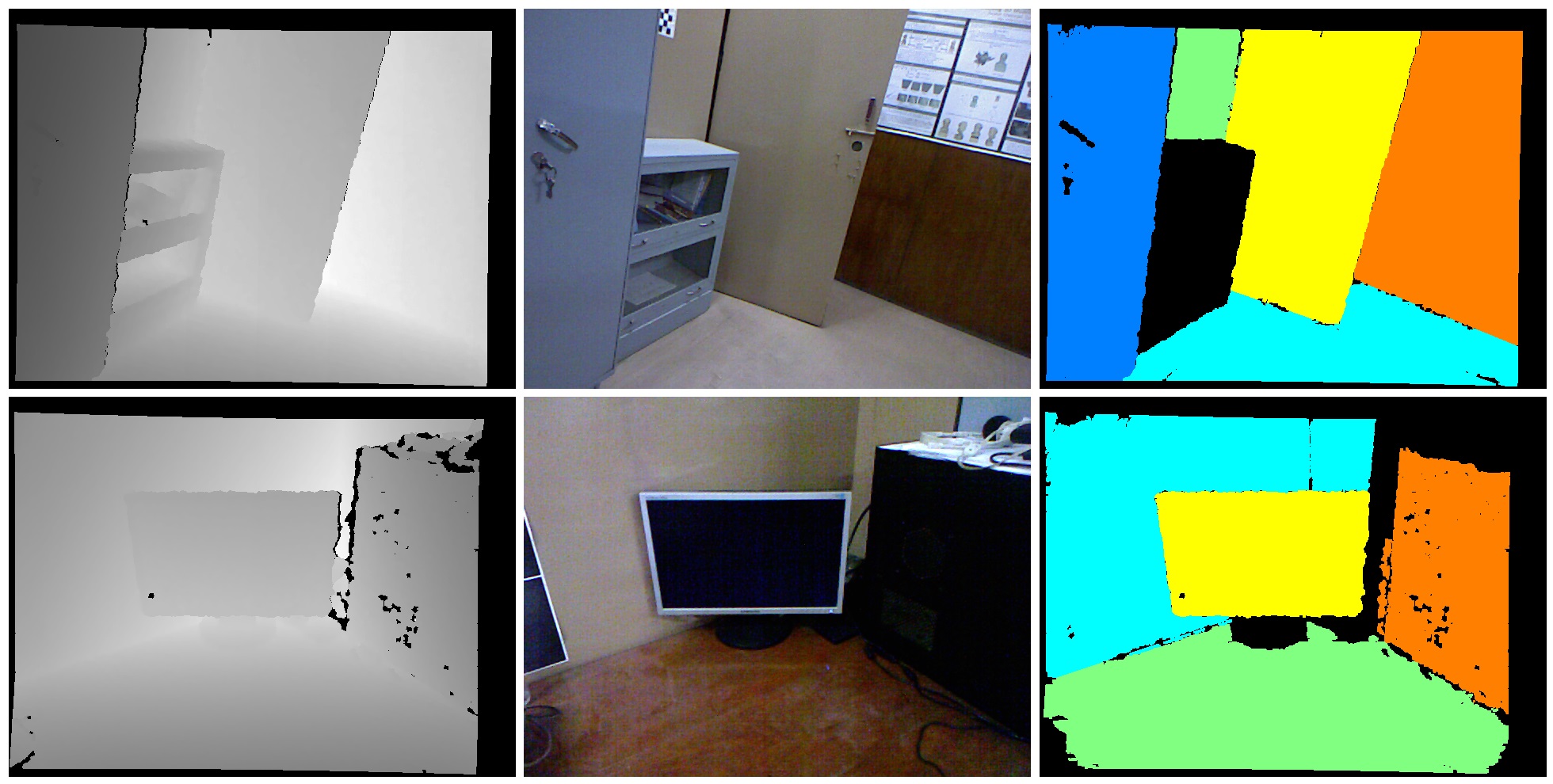}}\\
\subfigure[Plane extraction on instances from the RGB-D SLAM Dataset are shown as color coded regions overlaid on the corresponding RGB images. Regions identified as non-planar are shown in gray.]
{\includegraphics[width=.245\textwidth]{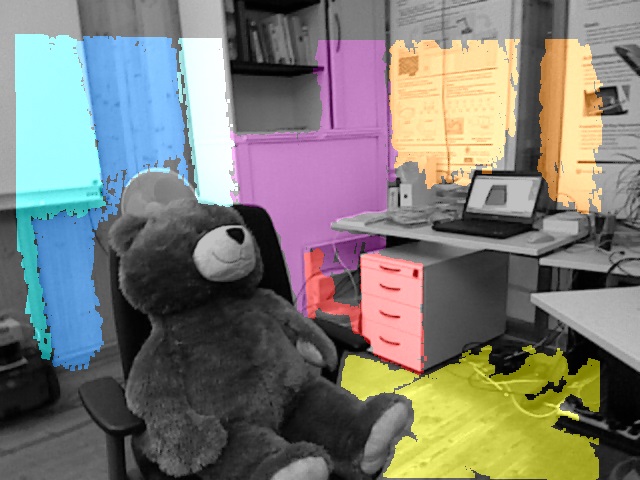}
\includegraphics[width=.245\textwidth]{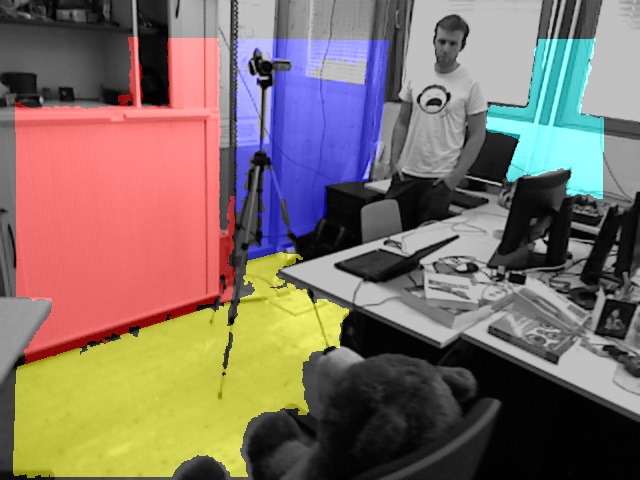}
\includegraphics[width=.245\textwidth]{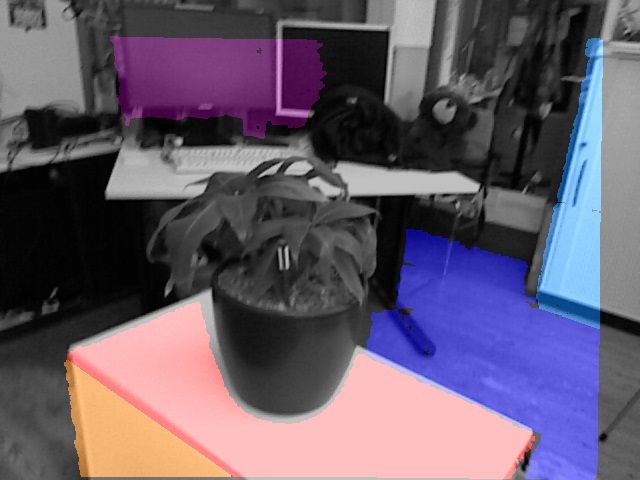}
\includegraphics[width=.245\textwidth]{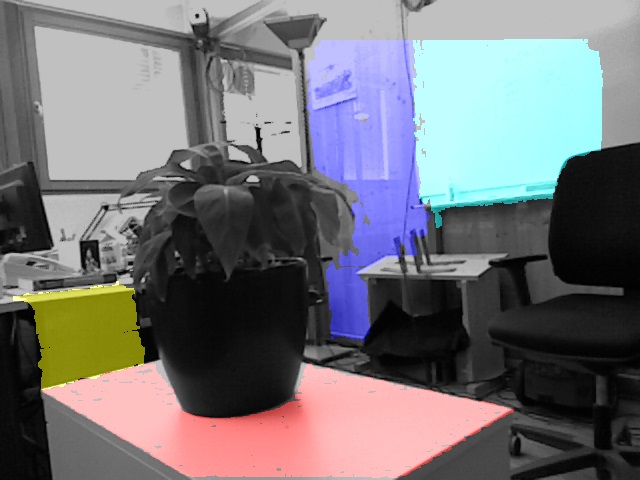}}
\caption{Results of applying our plane extraction method to different depth maps. Please see this figure in color.}
\label{Figure:ResultPlaneExtraction}
\end{figure*}

\begin{figure}
\centering
\includegraphics[width=.4\textwidth]{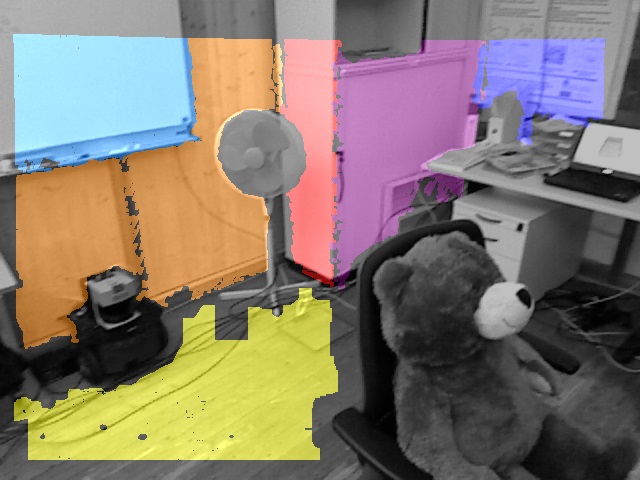}
\includegraphics[width=.4\textwidth]{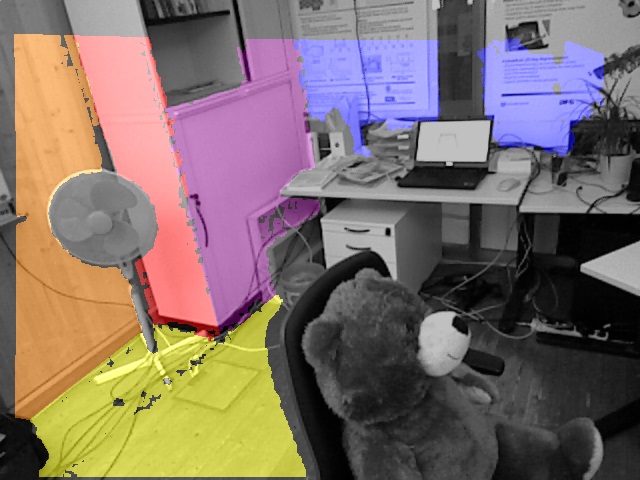}
\caption{Use of plane extraction for motion estimation. Plane extracted from two frames of teddy sequence using our method. Common planes are then manually matched and used for rotation estimation between these two frames. Please see text for details.}
\label{Fig:SLAM}
\end{figure}

In this subsection, we demonstrate some results of our disparity map based plane extraction algorithm. In Figure \ref{Figure:ResultPlaneExtraction}(a) we show our results on two real world scenes. In Figure \ref{Figure:ResultPlaneExtraction}(b) we show the result of our plane extraction algorithm applied to more complex scenes from the RGB-D SLAM dataset \cite{sturm12iros}. The plane extraction results of our method are depicted by overlaying color coded regions onto the corresponding RGB images. As can be seen, our approach effectively captures the significant planar regions present in the scene and this representation can be used for motion estimation in a SLAM context~\cite{TaguchiJRF12,DouGFF12,ataer2013tracking}. In Figure~\ref{Fig:SLAM}, we show the extracted planes as color coded regions on the corresponding RGB images for two depth frames of a SLAM sequence (frames 204 and 207  of the \texttt{freiburg1\textunderscore teddy} sequence of the RGBD-SLAM dataset (category 3D object reconstruction)). Using the corresponding analytic models of the matched or corresponding 3D planes, we estimate the 3D Euclidean motion between the two depth frames. For ground truth, we apply the point-to-plane ICP to estimate the Euclidean motion between the two depth frames.\footnote{Since the ground truth provided with this dataset is not synchronized in time with the raw data, we have chosen to use the results of ICP as ground truth for this experiment.}. The initial rotation between the two frames is $22.7^{\circ}$ whereas our rotation estimate bring the two frames to be as close as $2.36^{\circ}$. In other words, using our plane extraction method can provide a fast and reliable estimate of the motion of the depth camera which can be used to initialize 3D motion estimation in a SLAM framework. Apart from significantly speeding up the 3D motion estimation, by providing a good initial estimate for the relative motion between the two frames, our method can ensure that we are within the region of convergence for greedy algorithms such as ICP.
\section{Conclusion}\label{Sec:Conclusion}
In this paper we have studied the characteristics of the noise present in structured-light stereo based depth cameras. We derive a theoretical model to account for the standard deviation of noise present and then provide experimental validation of this noise model. In addition, we demonstrate the gains to be had by incorporating this noise model into three important applications, i.e. depth map denoising, volumetric scan merging and plane extraction. Extensive experimental results are presented for each of these applications that validate the utility of our noise model.

\section*{Acknowledgements}
The authors acknowledge the generous help of Shri Amrutbhai Modi, Tridip Suhrud and their colleagues at Gandhi Ashram at Sabarmati, Ahmedabad during the data acquisition for the results depicted in Figure~\ref{Figure:Mahatma}


\bibliography{paper}

\begin{thebibliography}{10}
\providecommand{\url}[1]{{#1}}
\providecommand{\urlprefix}{URL }
\expandafter\ifx\csname urlstyle\endcsname\relax
  \providecommand{\doi}[1]{DOI~\discretionary{}{}{}#1}\else
  \providecommand{\doi}{DOI~\discretionary{}{}{}\begingroup
  \urlstyle{rm}\Url}\fi

\bibitem{KinectROSSpec}
Kinect calibration.
\newblock \urlprefix\url{http://wiki.ros.org/kinect_calibration/technical}

\bibitem{KinectSDK}
Kinect for windows sdk v1.8.
\newblock
  \urlprefix\url{http://www.microsoft.com/en-in/download/details.aspx?id=40278}

\bibitem{KinectSpec}
Kinect for windows sensor components and specifications.
\newblock \urlprefix\url{http://msdn.microsoft.com/en-us/library/jj131033.aspx}

\bibitem{ataer2013tracking}
Ataer-Cansizoglu, E., Taguchi, Y., Ramalingam, S., Garaas, T.: Tracking an
  rgb-d camera using points and planes.
\newblock In: {ICCVW} Computer Vision Workshops, pp. 51--58. IEEE (2013)

\bibitem{CalibrationToolBox}
Bouguet, J.Y.: {Camera calibration toolbox for Matlab} (2008).
\newblock \urlprefix\url{http://www.vision.caltech.edu/bouguetj/calib_doc/}

\bibitem{camplani2013depth}
Camplani, M., Mantecon, T., Salgado, L.: Depth-color fusion strategy for 3-d
  scene modeling with kinect.
\newblock {IEEE} T. Cybernetics \textbf{43}(6), 1560--1571 (2013)

\bibitem{chatterjee2012pipeline}
Chatterjee, A., Jain, S., Govindu, V.M.: A pipeline for building 3d models
  using depth cameras.
\newblock In: The Eighth Indian Conference on Vision, Graphics and Image
  Processing, {ICVGIP}, p.~38 (2012)

\bibitem{ScalableReconstruction}
Chen, J., Bautembach, D., Izadi, S.: Scalable real-time volumetric surface
  reconstruction.
\newblock {ACM} Trans. Graph. \textbf{32}(4), 113 (2013)

\bibitem{ChenLL12}
Chen, L., Lin, H., Li, S.: Depth image enhancement for kinect using region
  growing and bilateral filter.
\newblock In: Proceedings of the 21st International Conference on Pattern
  Recognition, {ICPR}, pp. 3070--3073 (2012)

\bibitem{cho2013depth}
Cho, J.H., Ikehata, S., Yoo, H., Gelautz, M., Aizawa, K.: Depth map up-sampling
  using cost-volume filtering.
\newblock In: IVMSP Workshop, 2013 IEEE 11th, pp. 1--4. IEEE (2013)

\bibitem{CurlessLevoy}
Curless, B., Levoy, M.: A volumetric method for building complex models from
  range images.
\newblock In: {SIGGRAPH}, pp. 303--312 (1996)

\bibitem{DouGFF12}
Dou, M., Guan, L., Frahm, J., Fuchs, H.: Exploring high-level plane primitives
  for indoor 3d reconstruction with a hand-held {RGB-D} camera.
\newblock In: {ACCV} Computer Vision Workshops, pp. 94--108 (2012)

\bibitem{EndresHSCB14}
Endres, F., Hess, J., Sturm, J., Cremers, D., Burgard, W.: 3-{D} mapping with
  an {RGB}-{D} camera.
\newblock IEEE Transactions on Robotics \textbf{30}(1), 177--187 (2014)

\bibitem{engelhard2011real}
Engelhard, N., Endres, F., Hess, J., Sturm, J., Burgard, W.: Real-time 3d
  visual slam with a hand-held rgb-d camera.
\newblock In: Proc. of the RGB-D Workshop on 3D Perception in Robotics at the
  European Robotics Forum, vol. 180 (2011)

\bibitem{essmaeel2014comparative}
Essmaeel, K., Gallo, L., Damiani, E., De~Pietro, G., Dipanda, A.: Comparative
  evaluation of methods for filtering kinect depth data.
\newblock Multimedia Tools and Applications pp. 1--24 (2014)

\bibitem{garcia2013real}
Garcia, F., Aouada, D., Solignac, T., Mirbach, B., Ottersten, B.: Real-time
  depth enhancement by fusion for rgb-d cameras.
\newblock Computer Vision, IET \textbf{7}(5), 1--11 (2013)

\bibitem{VenuPooja2014}
Govindu, V.M., Pooja, A.: On averaging multiview relations for 3d scan
  registration.
\newblock Image Processing, IEEE Transactions on \textbf{23}(3), 1289--1302
  (2014)

\bibitem{hanhigh}
Han, Y., Lee, J., Kweon, I.: High quality shape from a single {RGB-D} image
  under uncalibrated natural illumination.
\newblock In: {IEEE} International Conference on Computer Vision, {ICCV}, pp.
  1617--1624 (2013)

\bibitem{HartleyZisserman}
Hartley, R., Zisserman, A.: Multiple view geometry in computer vision {(2.}
  ed.).
\newblock Cambridge University Press (2006)

\bibitem{RGBDMapping}
Henry, P., Krainin, M., Herbst, E., Ren, X., Fox, D.: Rgbd mapping: Using depth
  cameras for dense 3d modeling of indoor environments.
\newblock In: In RGB-D: Advanced Reasoning with Depth Cameras Workshop in
  conjunction with RSS (2010)

\bibitem{holz2012real}
Holz, D., Holzer, S., Rusu, R.B., Behnke, S.: Real-time plane segmentation
  using {RGB-D} cameras.
\newblock In: RoboCup 2011: Robot Soccer World Cup {XV}, pp. 306--317 (2011)

\bibitem{khoshelham2011accuracy}
Khoshelham, K.: Accuracy analysis of kinect depth data.
\newblock In: ISPRS workshop laser scanning, vol.~38, p. W12 (2011)

\bibitem{khoshelham2012accuracy}
Khoshelham, K., Elberink, S.O.: Accuracy and resolution of kinect depth data
  for indoor mapping applications.
\newblock Sensors \textbf{12}(2), 1437--1454 (2012)

\bibitem{KimCKA11}
Kim, S., Cho, W., Koschan, A., Abidi, M.A.: Depth map enhancement using
  adaptive steering kernel regression based on distance transform.
\newblock In: Advances in Visual Computing - 7th International Symposium,
  {ISVC}, pp. 291--300 (2011)

\bibitem{LiuGL12}
Liu, J., Gong, X., Liu, J.: Guided inpainting and filtering for kinect depth
  maps.
\newblock In: International Conference on Pattern Recognition, {ICPR}, pp.
  2055--2058 (2012)

\bibitem{LoWH13}
Lo, K., Wang, Y.F., Hua, K.: Joint trilateral filtering for depth map
  super-resolution.
\newblock In: Visual Communications and Image Processing, {VCIP}, pp. 1--6
  (2013)

\bibitem{lorensen1987marching}
Lorensen, W.E., Cline, H.E.: Marching cubes: {A} high resolution 3d surface
  construction algorithm.
\newblock In: Proceedings of the 14th Annual Conference on Computer Graphics
  and Interactive Techniques, {SIGGRAPH} 1987, pp. 163--169 (1987)

\bibitem{martinezkinect}
Martinez, M., Stiefelhagen, R.: Kinect unleashed: Getting control over high
  resolution depth maps.
\newblock In: Proceedings of the 13. {IAPR} International Conference on Machine
  Vision Applications, {MVA}, pp. 247--250 (2013)

\bibitem{Masuda02}
Masuda, T.: Registration and integration of multiple range images by matching
  signed distance fields for object shape modeling.
\newblock Computer Vision and Image Understanding \textbf{87}(1-3), 51--65
  (2002)

\bibitem{matyunin2011temporal}
Matyunin, S., Vatolin, D., Berdnikov, Y., Smirnov, M.: Temporal filtering for
  depth maps generated by kinect depth camera.
\newblock In: 3DTV Conference: The True Vision-Capture, Transmission and
  Display of 3D Video (3DTV-CON), pp. 1--4. IEEE (2011)

\bibitem{milani2012joint}
Milani, S., Calvagno, G.: Joint denoising and interpolation of depth maps for
  {MS} kinect sensors.
\newblock In: {IEEE} International Conference on Acoustics, Speech and Signal
  Processing, {ICASSP}, pp. 797--800 (2012)

\bibitem{montani1994modified}
Montani, C., Scateni, R., Scopigno, R.: A modified look-up table for implicit
  disambiguation of marching cubes.
\newblock The Visual Computer \textbf{10}(6), 353--355 (1994)

\bibitem{KinectFusion}
Newcombe, R.A., Izadi, S., Hilliges, O., Molyneaux, D., Kim, D., Davison, A.J.,
  Kohli, P., Shotton, J., Hodges, S., Fitzgibbon, A.W.: Kinectfusion: Real-time
  dense surface mapping and tracking.
\newblock In: {IEEE} International Symposium on Mixed and Augmented Reality,
  {ISMAR}, pp. 127--136 (2011)

\bibitem{nguyen2012modeling}
Nguyen, C.V., Izadi, S., Lovell, D.: Modeling kinect sensor noise for improved
  3d reconstruction and tracking.
\newblock In: International Conference on 3D Imaging, Modeling, Processing,
  Visualization {\&} Transmission, pp. 524--530 (2012)

\bibitem{HashingReconstruction}
Nie{\ss}ner, M., Zollh{\"{o}}fer, M., Izadi, S., Stamminger, M.: Real-time 3d
  reconstruction at scale using voxel hashing.
\newblock {ACM} Trans. Graph. \textbf{32}(6), 169 (2013)

\bibitem{pathak2010uncertainty}
Pathak, K., Vaskevicius, N., Birk, A.: Uncertainty analysis for optimum plane
  extraction from noisy 3d range-sensor point-clouds.
\newblock Intelligent Service Robotics \textbf{3}(1), 37--48 (2010)

\bibitem{poppinga2008fast}
Poppinga, J., Vaskevicius, N., Birk, A., Pathak, K.: Fast plane detection and
  polygonalization in noisy 3d range images.
\newblock In: IEEE/RSJ International Conference on Intelligent Robots and
  Systems, {IROS}, pp. 3378--3383. IEEE (2008)

\bibitem{qi2013structure}
Qi, F., Han, J., Wang, P., Shi, G., Li, F.: Structure guided fusion for depth
  map inpainting.
\newblock Pattern Recognition Letters \textbf{34}(1), 70--76 (2013)

\bibitem{MovingKinfu}
Roth, H., Vona, M.: Moving volume kinectfusion.
\newblock In: British Machine Vision Conference, {BMVC}, pp. 1--11 (2012)

\bibitem{EfficientICP}
Rusinkiewicz, S., Levoy, M.: Efficient variants of the {ICP} algorithm.
\newblock In: International Conference on 3D Digital Imaging and Modeling
  {3DIM}, pp. 145--152 (2001)

\bibitem{smisek20133d}
Smisek, J., Jancosek, M., Pajdla, T.: 3d with kinect.
\newblock In: Consumer Depth Cameras for Computer Vision, pp. 3--25. Springer
  (2013)

\bibitem{CPU3dMapping}
Steinbr{\"{u}}cker, F., Sturm, J., Cremers, D.: Volumetric 3d mapping in
  real-time on a {CPU}.
\newblock In: 2014 {IEEE} International Conference on Robotics and Automation,
  {ICRA}, pp. 2021--2028 (2014)

\bibitem{sturm12iros}
Sturm, J., Engelhard, N., Endres, F., Burgard, W., Cremers, D.: A benchmark for
  the evaluation of {RGB-D} {SLAM} systems.
\newblock In: {IEEE/RSJ} International Conference on Intelligent Robots and
  Systems, {IROS}, pp. 573--580 (2012)

\bibitem{TaguchiJRF12}
Taguchi, Y., Jian, Y., Ramalingam, S., Feng, C.: {SLAM} using both points and
  planes for hand-held 3d sensors.
\newblock In: {IEEE} International Symposium on Mixed and Augmented Reality,
  {ISMAR}, pp. 321--322 (2012)

\bibitem{BilateralFilter}
Tomasi, C., Manduchi, R.: Bilateral filtering for gray and color images.
\newblock In: International Conference on Computer Vision {ICCV}, pp. 839--846
  (1998)

\bibitem{turk1994zippered}
Turk, G., Levoy, M.: Zippered polygon meshes from range images.
\newblock In: Proceedings of the 21th Annual Conference on Computer Graphics
  and Interactive Techniques, {SIGGRAPH}, pp. 311--318 (1994)

\bibitem{WangZPQ14}
Wang, Y., Zhong, F., Peng, Q., Qin, X.: Depth map enhancement based on color
  and depth consistency.
\newblock The Visual Computer \textbf{30}(10), 1157--1168 (2014)

\bibitem{Kintinuous}
Whelan, T., Kaess, M., Fallon, M., Johannsson, H., Leonard, J., McDonald, J.:
  Kintinuous: Spatially extended {K}inect{F}usion.
\newblock In: RSS Workshop on RGB-D: Advanced Reasoning with Depth Cameras.
  Sydney, Australia (2012)

\bibitem{yu2013shading}
Yu, L., Yeung, S.K., Tai, Y., Lin, S.: Shading-based shape refinement of
  {RGB-D} images.
\newblock In: {IEEE} Conference on Computer Vision and Pattern Recognition, pp.
  1415--1422 (2013)

\bibitem{EfficientKinfu}
Zeng, M., Zhao, F., Zheng, J., Liu, X.: A memory-efficient kinectfusion using
  octree.
\newblock In: Computational Visual Media - First International Conference,
  {CVM}, pp. 234--241 (2012)

\bibitem{ZhaoTFTCC13}
Zhao, M., Tan, F., Fu, C., Tang, C., Cai, J., Cham, T.: High-quality kinect
  depth filtering for real-time 3d telepresence.
\newblock In: {IEEE} International Conference on Multimedia and Expo, {ICME},
  pp. 1--6 (2013)

\bibitem{InterestPoints}
Zhou, Q., Koltun, V.: Dense scene reconstruction with points of interest.
\newblock {ACM} Trans. Graph. \textbf{32}(4), 112 (2013)

\bibitem{ElasticFragments}
Zhou, Q., Miller, S., Koltun, V.: Elastic fragments for dense scene
  reconstruction.
\newblock In: {IEEE} International Conference on Computer Vision, {ICCV}, pp.
  473--480 (2013)

\end{thebibliography}

\end{document}